\documentclass[10pt,journal,cspaper,compsoc]{IEEEtran}
\ifCLASSOPTIONcompsoc
  \usepackage[nospace,nocompress]{cite}
\else
  \usepackage{cite}
\fi
\ifCLASSINFOpdf
\else
\fi
\usepackage{times}
\usepackage{epsfig}
\usepackage{graphicx}
\usepackage{amsmath}
\usepackage{amssymb}
\usepackage{multirow}
\usepackage[algoruled,vlined,linesnumbered]{algorithm2e}
\usepackage{rotating}
\usepackage{bbm}
\usepackage{bm}
\usepackage{color}
\usepackage{url}
\usepackage{threeparttable}
\usepackage{algorithmic}
\usepackage{subfigure}

\renewcommand{\algorithmicrequire}{\textbf{Input:}}   
\renewcommand{\algorithmicensure}{\textbf{Output:}}  
\DeclareMathAlphabet{\mathpzc}{OT1}{pzc}{m}{it} 

\DeclareMathOperator*{\argmin}{argmin}

\newcommand{\change}[1]{{\color{black}#1}}

\newcommand{\changed}[1]{\textcolor{black}{#1}}

\begin{document}
%
\title{Human-In-The-Loop Person Re-Identification}
%
%
%
%

\author{Hanxiao~Wang,
        ~Shaogang~Gong,
        ~Xiatian~Zhu,
        and Tao Xiang
\IEEEcompsocitemizethanks{\IEEEcompsocthanksitem 
	Hanxiao Wang is with the Electrical and Computer Engineering Department,
	Boston University, US.
	Email: hxw@bu.edu. 
	Shaogang Gong and Tao Xiang are with the School
	of Electronic Engineering and Computer Science, Queen Mary University of London, UK.
	Email: s.gong@qmul.ac.uk, t.xiang@qmul.ac.uk. 
	Xiatian Zhu is with Vision Semantics Ltd., 
	London, UK.
	Email: eddy@visionsemantics.com.
}
\thanks{}}

%
%

\markboth{}%
{H. Wang \MakeLowercase{\textit{et al.}}:  Human-In-The-Loop Person Re-identification}
%



\IEEEtitleabstractindextext{%
\begin{abstract}
Current person re-identification (re-id) methods assume that (1) pre-labelled training data is available for every camera pair, (2) the gallery size is moderate in model deployment. However, both assumptions are invalid in real-world applications where camera network and gallery size increase dramatically. Under such more realistic conditions, human involvement is often inevitable to verify the results generated by an automatic computer algorithm. In this work, rather than proposing another fully-automated and yet unrealistic re-id model, we introduce a semi-automatic re-id solution. Our goal is to minimise human efforts spent in re-id deployments, while maximally drive up re-id performance. Specifically, a {\em hybrid} human-computer re-id model based on Human Verification Incremental Learning (HVIL) is formulated which does not require any pre-labelled training data, therefore scalable to new camera pairs; Moreover, this HVIL model learns cumulatively from human feedback to provide an instant improvement to re-id ranking of each probe on-the-fly, thus scalable to large gallery sizes. We further formulate a Regularised Metric Ensemble Learning (RMEL) model to combine a series of incrementally learned HVIL models into a single ensemble model to be used when human feedback becomes unavailable. We conduct extensive comparative evaluations on three benchmark datasets (CUHK03, Market-1501, and VIPeR) to demonstrate the advantages of the proposed HVIL re-id model over state-of-the-art conventional human-out-of-the-loop re-id methods and contemporary human-in-the-loop competitors. 
\end{abstract}

}

\maketitle

\IEEEdisplaynontitleabstractindextext

%
\IEEEpeerreviewmaketitle


\section{Introduction}
\label{sec:introduction}

%
%
%
%

\begin{figure} 
	\subfigure[\scriptsize Cross-view appearance variations]{
		\includegraphics[height=0.23\linewidth]{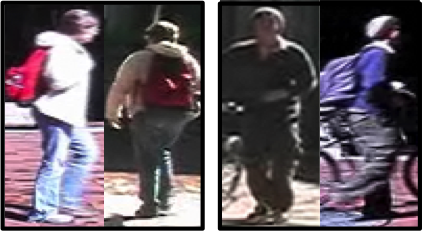}}
	\subfigure[\scriptsize Similar appearance among different people]{
		\includegraphics[height=0.23\linewidth]{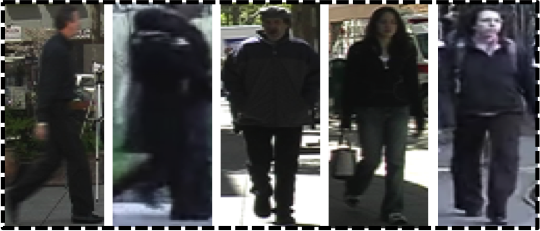}}
	\vskip -.3cm
	\caption{
		Person re-identification Challenges.
		(a) A significant
        visual appearance change of the same person across
        camera views. 
        (b) Strong appearance similarities among different people.
	}
	\label{fig:reid_challenges}
	\vspace{-.1cm}
\end{figure}

\IEEEPARstart{P}{erson}
re-identification (re-id) is the problem of matching people
across non-overlapping camera views distributed in
open spaces at different locations, typically achieved 
by matching detected bounding box images of people~\cite{gong2014person}.
This is an inherently challenging problem due to the potentially dramatic 
visual appearance changes caused by uncontrolled variations in human pose
and unknown viewing conditions on illumination, occlusion, and
background clutter (Fig.~\ref{fig:reid_challenges}). 
%
%
\changed{A re-id model is required to differentiate images of different categories (persons) with
similar appearances, which can be considered as solving a fine-grained visual categorisation problem \cite{wang2014learning,cui2016fine}, whilst also able to recognise a same category (person) with visually dissimilar appearances.}
Unlike conventional
biometrics identification problems, e.g. face recognition, a person
re-id model has {\em no labelled training data on target classes},
i.e. similar to a {\em zero-shot learning} problem \cite{kodirov2015unsupervised} that requires the model
to perform inherently transfer learning between a training population (seen)
and a target population (unseen). Moreover, person re-id requires
implicitly a model to perform {\em cross-domain transfer learning} \cite{pan2009transfer_survey} if each
camera view is considered as a specific domain of potentially
significant difference to other domains (views). This is
more difficult than a standard zero-shot learning problem.

\begin{figure*} [t] 
	\includegraphics[width=1\linewidth]{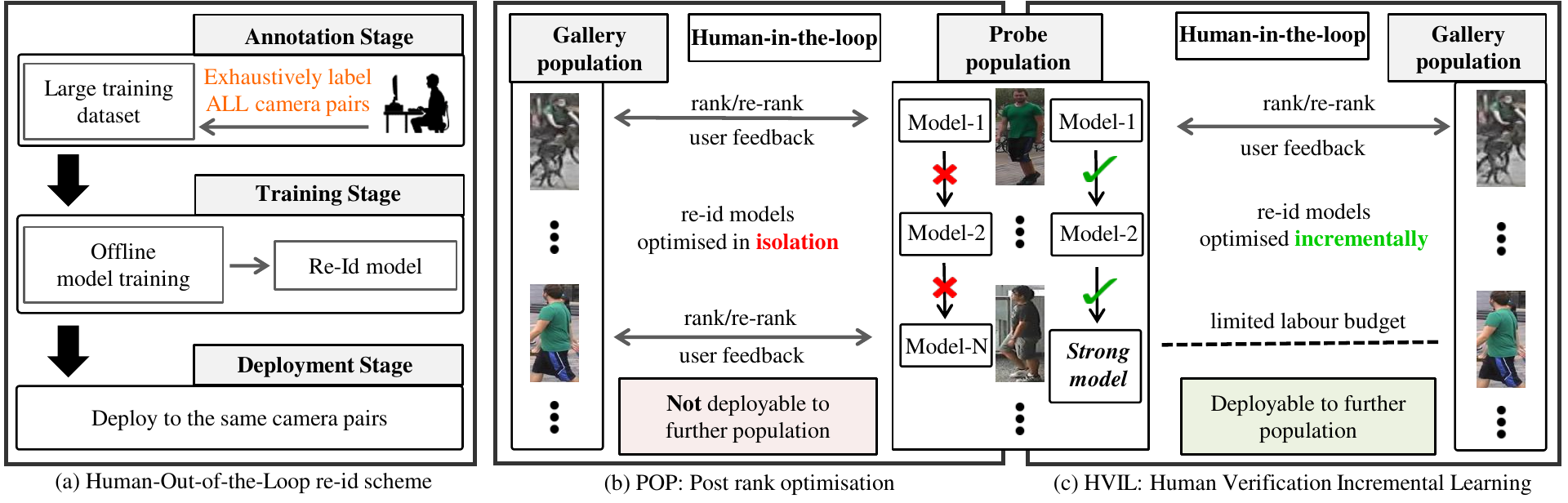}
	\vskip -.3cm
	\caption{
		Illustration of two person re-id schemes.
		{\bf (a)} The conventional {\em Human-Out-of-the-Loop} (HOL) re-id scheme
		requires exhaustive pre-labelled training data collection
		for supervised \change{offline} model learning. The learned model is assumed
		sufficiently \change{generalisable} and then deployed to perform fully automated person 
		re-id tasks without human in the loop.
		{\bf (b)} POP~\cite{liu2013pop}: A recent 
		{\em Human-In-the-Loop} (HIL) re-id approach which optimises
		probe-specific models in isolation from human feedback verifications in the deployment time.
		All probe people requires human in the loop. 
		{\bf (c)} HVIL: The proposed new incremental HIL re-id model capable of not only progressively learning
		a generalised model from human verifications across all probed people 
		while carrying out the HIL re-id tasks, but also performing the HOL re-id tasks
		when human effort becomes unavailable.
	}
	\label{fig:reid_diagram}
	\vspace{-.1cm}
\end{figure*}

Current re-id methods are
dominated by supervised learning techniques
\cite{gong2014person,PCCA_CVPR12,KISSME_CVPR12,PRD_PAMI13,CVPR13LFDA,Zhao_MidLevel_2014a,wang2014person,wang2016pami,Li_DeepReID_2014b,xiong2014person,liao2015person,Liao_2015_ICCV},
which typically employ a ``{\em train-once-and-deploy}'' scheme
(Fig.~\ref{fig:reid_diagram}(a)). 
That is, a pre-labelled training
dataset of pairwise true- and false-matching identities (training
population) is collected by human annotators for every pair of cameras through 
manually examining a vast pool of image/video data. This training
dataset is used to train an offline re-id model. 
It is tacitly assumed by most that
such a trained model can be deployed plausibly 
as a {\em fully automated}
solution to re-identify target (unseen during model training) person
images at test time, 
without any human assistance nor model adaptation.
%
Based on this assumption, the re-id community has witnessed 
ever-increased matching accuracies on
increasingly larger sized benchmarks of more training identity classes
over the past two years. For instance, the CUHK03
benchmark~\cite{Li_DeepReID_2014b} contains 13,164 images of 1,360 
identities, of which 1,260 are used for training and 100 for testing,
significantly larger than the earlier VIPeR \cite{VIPeR} (1,264 images
of 632 people with 316 for training)
and iLIDS~\cite{iLIDS_BMVC09} (476 images for 119 people with 69 for training). The
state-of-the-art Rank-1 accuracy on CUHK03 has exceeded $70\%$
\cite{xiao2016learning}, tripling the best performance reported only two years
ago \cite{Li_DeepReID_2014b}.

Despite such rapid progresses, current automatic re-id solutions
remain ill-suited for a practical deployment. This is because:
{\bf (1)} A manually pre-labelled 
pairwise training data set {\em for every
camera pair} does not exist, due to either being prohibitively expensive to collect in the real-world as there are a quadratic
number of camera pairs, 
or nonexistence of sufficiently large number of training people reappearing in every pair
of camera views. 
{\bf (2)} Assuming the size of the training population is either significantly greater or no less than that of 
the target population is unrealistic. For instance, the standard CUHK03 benchmark test defines
the training set having paired images of 1,260 people from six different 
camera views (on average 4.8 image samples per person per camera view), whilst
the test set having only 100 identities each with a single image. 
The test population is thus 10 times smaller than
the training population, with approximately 50 times less
images. In practice, any deployment gallery size (test population) is
almost always much greater than any labelled training data size even if
such training data were available. 
In a public space such as an underground station, there
are easily thousands of people passing through a camera view
every hour \cite{waterloo}, with a typical gallery population size of over 10,000 per day. 
%
We observed
on the CUHK03 dataset that, only a 10-fold increase in gallery size leads to
a 10-fold decrease in re-id Rank-1 performance, leading to a
single-digit Rank-1 score, even when the
state-of-the-art re-id models were trained from sufficiently sized labelled data. Given
such low Rank-1 scores, in practice human operators (users) would
still be required to verify
any true match of a probe from an automatically generated ranking list.
%
%
%

\begin{figure*} [t] 
	\centering
	\subfigure[Exhuastive labelling]{
		\includegraphics[height=0.138\linewidth]{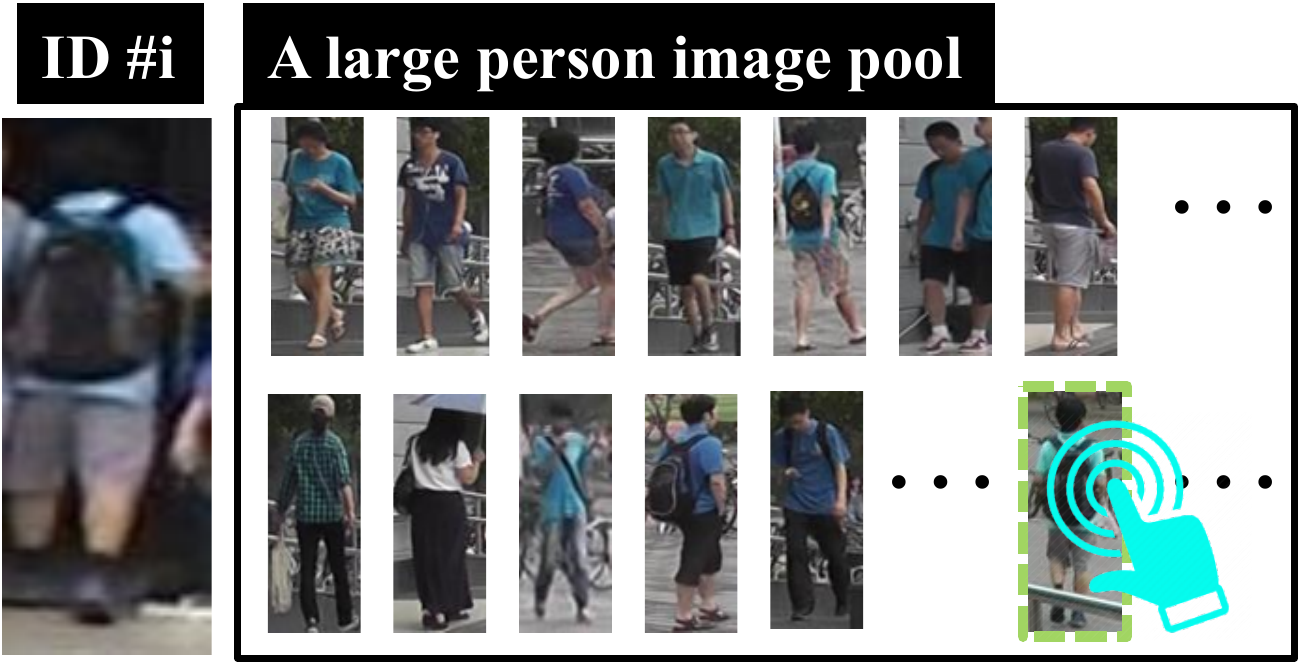}
	}
	\subfigure[True or false pairwise labelling]{
		\includegraphics[height=0.138\linewidth]{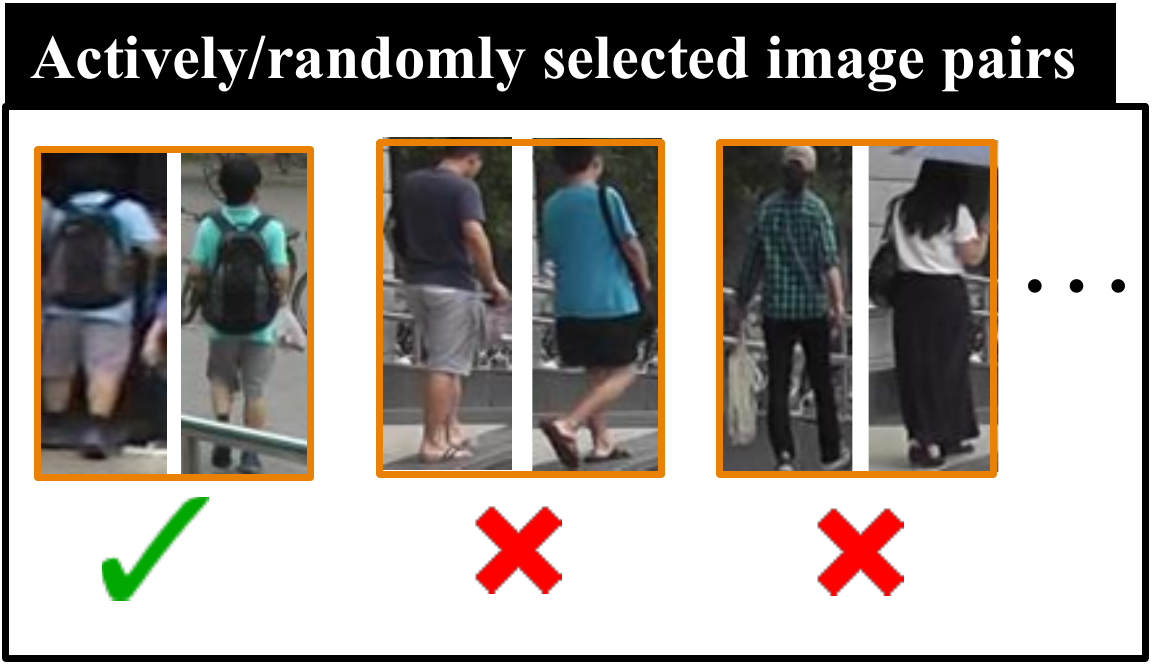}
	}
	\subfigure[Attribute labelling]{
		\includegraphics[height=0.138\linewidth]{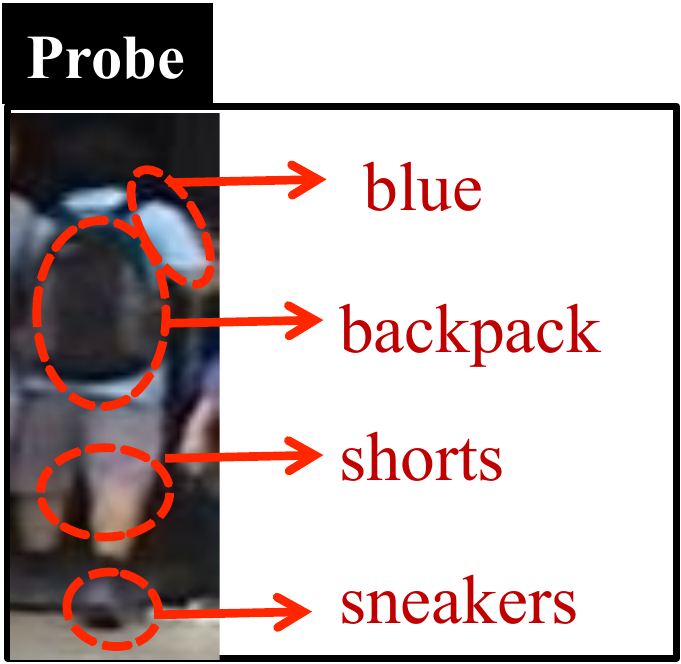}
	}
	\subfigure[Top ranks labelling (true match, negative)]{
		\includegraphics[height=0.138\linewidth]{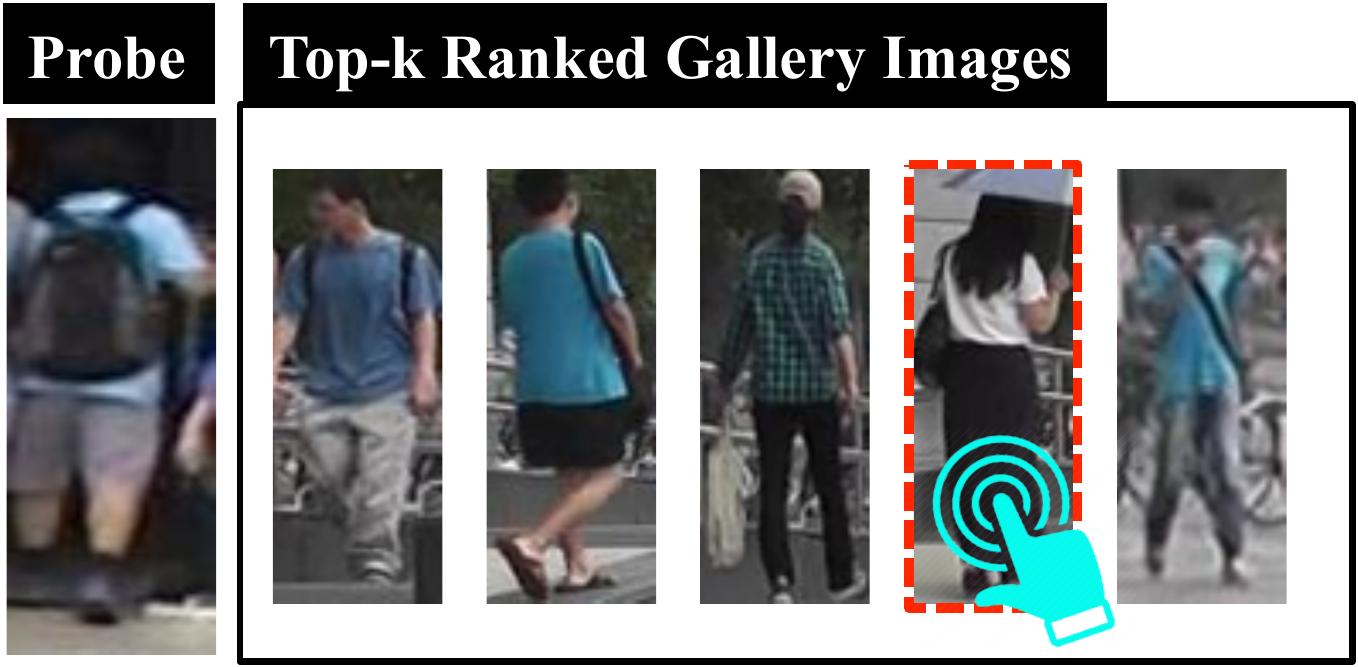}
	}
	\vskip -.2cm
	\caption{
		Different human labelling processes are employed in person re-id model training and deployment.
		{\bf (a)} Large size offline labelling of cross-view
                positive- and negative-pairs of training data with
                identity labels \cite{Anton_2015_CoRR,Liao_2015_ICCV,wang2016highly,zhang2016learning}. 
		{\bf (b)} Selective or random sampling of person image
                pairs for human verification in either model
                training~\cite{martinel2016temporal} or deployment~\cite{das2015active}. 
		{\bf (c)} Fine-grained attribute labelling in either training~\cite{su2016deep} or deployment~\cite{das2015active}.
		{\bf (d)} True match verification among the top ranked sub-list
                in model
                deployment~\cite{Hirzer_2011_SCIA,liu2013pop,wang2016human}, or
                verification of both visually dissimilar and similar
                wrong matches in top ranks (strong/hard and weak
                negative mining) in model deployment~\cite{liu2013pop,wang2016human}.
	}
	\vspace{-.1cm}
	\label{fig:reid_feedbacks}
\end{figure*}

In this work, we explore an alternative approach to person re-id by
formulating a hybrid human-computer learning paradigm with humans in the model
matching loop (Fig.~\ref{fig:reid_diagram}(c)).
%
\change{We call this semi-automated scheme {\it Human-In-the-Loop} (HIL)
	re-id, designed to optimise re-id performance given a small number of human
	verification feedback 
	and a larger-sized test population, as compared
	to the conventional {\it Human-Out-of-the-Loop} (HOL) re-id models that are
	mostly designed to optimise re-id given a larger-sized pre-labelled training
	data and a small-sized test population.}
%
%
%
%
This HIL re-id scheme has three significant advantages 
over the conventional HOL models: 
{\bf (1)} {\em Less human labelling effort}: 
HIL re-id requires much less human labelling effort, since 
it does not necessarily require
the expensive construction of a pre-labelled training set.
More importantly, it prioritises directly the human labour effort on each
given re-id task in deployment, 
rather than optimising the model learning error on
an independent training set. 
More specifically, the number of feedback from human verification is typically in {\em tens} as compared to
{\em thousands} of offline pre-labelled training data required by HOL methods. 
%
{\bf (2)} {\em Model transfer learning}:
Our HIL model is able to achieve greater transferability with better re-id
performance in test domains.
This is because a HIL model focuses 
on re-id matching optimisation directly in the deployment
gallery population, rather than learning 
a distance metric from a separate training
set and {\em assuming} its blind transferability to independent (unseen) test data.
It enables a human operator to
interactively validate model matching results for each re-id task
and inform on model mistakes (similar in spirit to hard negative mining). 
{\bf (3)} {\em Reinforcing visual consistency}:
As computer vision algorithms
are intrinsically very different from the human visual system, 
a re-id model can make mistakes that generate ``unexpected''
(visually inconsistent) re-id ranking results,
readily identifiable by a human observer.
By learning directly from the inconsistency between a
computer vision model and human observation, 
a HIL re-id model is guided to maximise visually more consistent
ranking lists favoured by human observations, and thus
\change{rendering the learned model more discriminative and capable of avoiding future mistakes that seem insensible to human observation.}
The main {\bf contribution} of this work is a novel HIL re-id model that
enables a user to re-identify rapidly a given probe person image after only 
a handful of feedback verifications even when the search gallery size is large.  
More specifically, a {\em Human Verification Incremental Learning} 
(HVIL) model (Fig.~\ref{fig:reid_diagram}(c)) is formulated
to {\em simultaneously} minimise human-in-the-loop feedback and 
maximise model re-id accuracy by incorporating:
{\bf (1)} \textit{Sparse feedback -} 
HVIL allows for easier human feedback on a few dissimilar matching results without the
need for exhaustive eyeball search of true/false in the entire rank list. 
It aims to rectify rapidly \change{unexpected} model mistakes by focusing {\em only} on
minimising visually obvious errors (hard negatives) identified by human
observation. This is reminiscent to learning by hard negative mining
but {\em with} human in the loop, so to improve model learning
with less training data. 
{\bf (2)} \textit{Immediate benefit -} 
HVIL introduces a new online incremental distance metric learning
algorithm, which enables real-time model response to human feedback by rapidly
presenting a freshly optimised ranking list for further human
feedback, quickly leading to identifying a true match. 
{\bf (3)} \textit{The older the wiser -}
HVIL is updated cumulatively on-the-fly utilising multiple user
feedback per probe, with incremental model optimisation for each new probe
given what have been learned from all previous probes.
{\bf (4)} \textit{A strong ensemble model -} 
An additional Regularised Metric Ensemble Learning
(RMEL) model is introduced by taking all the incrementally
optimised per-probe models as a set of ``weak'' models 
\cite{Schapire_1990_JML,Amit_1997_JNC} and constructing a ``strong''
ensemble model for performing HOL re-id tasks when human feedback becomes unavailable.
Extensive comparative experiments on three benchmark datasets 
(CUHK03 \cite{Li_DeepReID_2014b}, Market-1501 \cite{zheng2015scalable}, and VIPeR \cite{VIPeR}) 
demonstrate that this HVIL model outperforms the state-of-the-art methods
for both the proposed new HIL and the conventional HOL re-id deployments.

\begin{figure*}  
	\centering
	\includegraphics[width=0.95\linewidth]{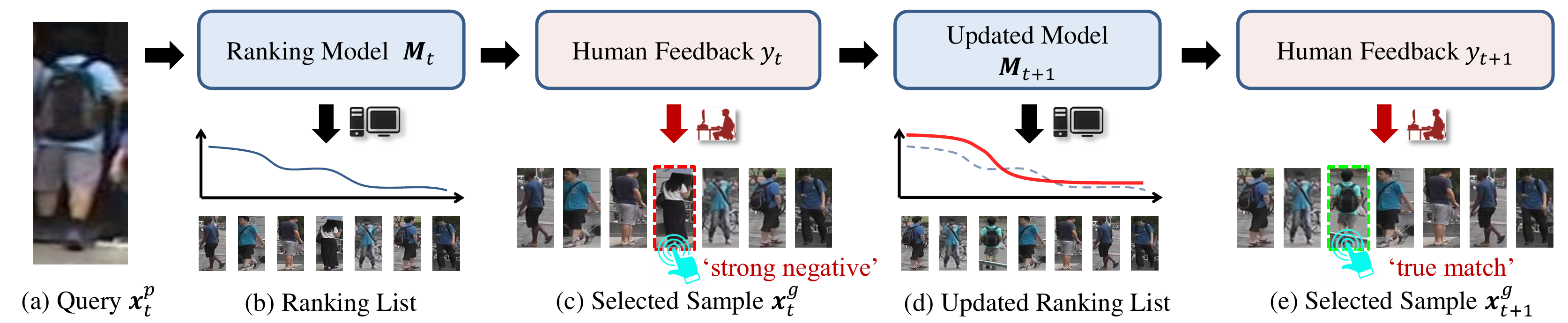}
	\vskip -0.3cm
	\caption{
		Visualisation of the proposed Human-In-the-Loop person re-id procedure. 
	}
	\vspace{-.1cm}
	\label{fig:model_flow}
\end{figure*}

\section{Related Work}
\label{sec:related_work}
\noindent {\bf Person Re-Identification } 
Current best performing person re-id methods are fully supervised and
require a large number of pre-labelled training data (Fig.~\ref{fig:reid_feedbacks}(a)) from every camera pair 
for building camera-pair specific distance metric models
\cite{REIDchallenge,PCCA_CVPR12,KISSME_CVPR12,PRD_PAMI13,CVPR13LFDA,Zhao_MidLevel_2014a,xiong2014person,wang2014person,wang2016pami,ding2015deep,Anton_2015_CoRR,liao2015person,Liao_2015_ICCV,wang2016highly,GOG,zhang2016learning}. 
Their usability and scalability are inherently limited in real-world applications 
especially with large camera networks. 
This problem becomes more acute for the more recent data-hungry
deep learning based
methods~\cite{UstinovaGL15,ShiZLLYL15,ding2015deep,Ahmed2015CVPR,Li_DeepReID_2014b,chen2016similarity,Gated_SCNN,LSSCDL,ding2015deep,wangjoint,xiao2016learning} which need
more labelled training data to function. 
To relax this need for   
intensive labelling, existing attempts include 
semi-supervised \cite{CVPR14semiReid,kodirovdictionary},
unsupervised \cite{SDALF_CVPR10,SalienceReId_CVPR13,hanxiao2014GTS,ElyorECCV16}, 
and transfer learning~\cite{Transfer_Ryan_ARTEMIS13,Transfer_ICCV13,wang2015cross,ma2015cross}.
However, all of these strategies are weak in performance compared to
fully supervised learning - without labelled data, they are unable to
learn strong discriminative information for cross-view re-identification.
In general, all existing methods are aimed for automated HOL re-id deployment, thus
suffering from dramatic performance degradation given a small size training
population, even with the best state-of-the-art supervised
method~\cite{liao2015person,wang2016highly,zhang2016learning,Liao_2015_ICCV,GOG,xiong2014person},
as can be seen in our evaluations. 
In contrast, the proposed HVIL model learns
interactively from human online feedback equivalent to a smaller number
of {\em selective} labelling of negative-pair data on-the-fly, therefore costing less
human ``labelling effort''. This HVIL approach to re-id yields
superior re-id matching accuracy than 
the state-of-the-art conventional supervised learning models, especially when
the deployment gallery size becomes larger. 

\vspace{0.1cm}
\noindent {\bf Hard Negative Mining }
As an effective scheme for improving model learning,
negative mining has been extensively exploited in tackling computer vision problems,
e.g. pedestrian detection \cite{felzenszwalb2010object},
face recognition \cite{schroff2015facenet},
image categorisation \cite{oh2016deep,wang2014learning,huang2016local},
and unsupervised visual representation learning \cite{wang2015unsupervised}.
Although hard negative samples are often collected automatically by
deploying an up-to-date model with ground-truth labels, human
verification can provide an alternative approach to hard negative mining when ground-truth labels
are unavailable~\cite{cui2016fine}. 
In contrast to \cite{cui2016fine}, HVIL has three significant advantages:
(1) HVIL does not require pre-collected labelled training data. This
is both necessary and critical for initialising the deep model of~\cite{cui2016fine}. 
(2) HVIL requires only a handful of hard negative verifications from the top ranks,
whereas a much larger number of human feedback is needed by
\cite{cui2016fine} in order to have sufficient data for a deep model
learning update. This makes HVIL much cheaper to be updated thus more scalable.
(3) Due to the inherently costly training time for a deep model, 
it is infeasible for \cite{cui2016fine} to perform on-the-fly
immediate model output update in responding to human feedback, losing
the model effectiveness and cognitive benefit to the human in the loop. 
For person re-id, the recent POP model~\cite{liu2013pop} also exploits
hard negative mining by human verification. In comparison, the
proposed HVIL human feedback protocol is simpler than that of POP, so that
a user {\em only} focuses on the most salient negatives. This is more
cost-effective than POP, as shown in our evaluation (Sec. \ref{sec_eval:human}). 

\vspace{0.1cm}
\noindent {\bf Online Learning } For learning from human feedback
on-the-fly, the sequential and iterative model update
in real-time poses an online learning problem
\cite{rosenblatt1958perceptron}. 
In general, online learning algorithms have been widely applied in
computer vision, especially for large-scale problems such as image
retrieval~\cite{wan2015online,piras2013passive},
ranking~\cite{Chechik_2010_JMLR}, and
classification~\cite{davis2007information}. 
\change{Algorithmically}, the proposed HVIL approach shares the principles of
Passive-Aggressive (PA) online learning~\cite{crammer2006online}, where
the learning objective has two terms: (1) A passive term that enforces the
consecutively learned model parameter values to be close to each other
at each time step, so to preserve the information learned from
past data. (2) An aggressive term that encourages the model to
incorporate information from new data. This enables the HVIL model to
\change{progressively} cumulate knowledge from previous human feedback, therefore
preserving human annotation effort and adapting to new data. Our
experiments demonstrate its effectiveness (Sec.~\ref{sec:model_feedback}).

\vspace{0.1cm}
\noindent {\bf Interactive Learning } 
Interactive model learning with human-in-the-loop is attractive for two reasons: 
(1) It provides a user with tools that can 
significantly alleviate or even eliminate the need for careful
preparation of large-sized training data. (2) It allows to reduce the
human labelling effort by exploiting a model's capacity interactively.
Human-computer interactive models have been considered in
image segmentation \cite{singaraju2008interactive,rother2004grabcut},
object recognition \cite{branson2010visual,wah2011multiclass},
semi-supervised clustering \cite{lad2014interactively}
and object counting \cite{arteta2014interactive}.
In addition, relevance feedback \cite{zhou2003relevance,lin2015effect,xu2011efficient}
and active learning \cite{settles2010active,hospedales2012unifying} are also related
to a similar idea of exploiting human feedback to improve model learning.
The former has been exploited for interactive image retrieval 
where human feedback to search results are used to refine a query. 
The latter aims to reduce the human labelling 
effort by active sample selection for model training.
In active learning, knowledge cumulation during model deployment is
not considered, and some offline pre-labelled data are
typically needed for model initialisation.

\vspace{0.1cm}
\noindent {\bf Human-In-the-Loop Person Re-Id } 
A small number of HIL re-id methods have been proposed recently.
Abir et al.~\cite{das2015active} (Fig.~\ref{fig:reid_feedbacks}(b,c))
exploited human-in-the-loop verification to expand
their multi-class based re-id model.
Compared to HVIL, their method requires a pre-labelled training set for model initialisation.
Another limitation is that such a model cannot generalise to new
person classes re-id when human effort becomes unavailable. 
%
Hirzer et al.~\cite{Hirzer_2011_SCIA} (Fig.~\ref{fig:reid_feedbacks}(d))
considered a form of human feedback which is ill-posed in practice: It only allows a user to 
verify whether a {\em true} match is within the top-$k$ ranking list.
This limits significantly the effectiveness of human feedback and can
waste expensive human labour when a true match cannot be found in
the top-$k$ ranks, which is rather typical for a re-id model
trained by small-sized training data and deployed to a larger-size test
gallery population. 
More recently, 
Liu et al.~\cite{liu2013pop}  proposed the POP model (Fig.~\ref{fig:reid_diagram}(d) and Fig. \ref{fig:reid_feedbacks}(d)), which
allows a user to identify correct matches
more rapidly and accurately by accommodating more flexible human feedback.
However, POP requires to perform label propagation on an
affinity graph over all gallery samples.
This makes it poor for large gallery sizes (Sec.~\ref{sec:exp}). 
Moreover, all existing HIL re-id
models~\cite{das2015active,Hirzer_2011_SCIA,liu2013pop} do not benefit
from cumulative learning, i.e. they treat each probe re-id as an
independent modelling or retrieval task; therefore the process of model
learning for re-id each probe does not benefit learning the models for other
probes. This lack of improving model-learning cumulatively from increased
human feedback is both suboptimal and disengaging the human in the loop. 
In contrast, the proposed HVIL re-id framework
(Fig.~\ref{fig:reid_diagram}(c) and Fig. \ref{fig:reid_feedbacks}(d)) enables incremental model improvement
from cumulative human feedback thus maximising and encouraging human-machine interaction.
Moreover, the proposed RMEL ensemble model further benefits from
previous human verification effort to 
enable conventional HOL re-id tasks when human feedback is no long available. 
An earlier and preliminary version of this work was presented in \cite{wang2016human}. Comparing to \cite{wang2016human}, this paper provides more detailed discussions,
additional theoretical analysis, and more extensive 
evaluations.

\section{Human-In-the-Loop Incremental Learning}
\label{sec:method_incre}

\subsection{Problem Formulation}
\label{problem def}


Let a person image be denoted by a feature vector $\bm{x} \in
\mathbb{R}^d$. The {\em Human-In-the-Loop (HIL) re-id} problem is formulated
as: {\bf (1)} For each image $\bm{x}^p$ in a probe set 
$\mathcal{P} = \{\bm{x}_i^p\}_{i=1}^{N_p}$ (Fig. \ref{fig:model_flow}(a)), 
$\bm{x}^p$ is matched against a gallery set $\mathcal{G} =
\{\bm{x}_i^g\}_{i=1}^{N_g}$ and an initial ranking list for all gallery images is 
generated by a re-id ranking function $f(\cdot): \mathbb{R}^d \rightarrow \mathbb{R}$, 
according to ranking scores $f_{\bm{x}^p}(\bm{x}^g_i)$ (Fig. \ref{fig:model_flow}(b)). 
{\bf (2)} A human operator (user) browses the gallery ranking list
to verify the existence and the rank of any true match for $\bm{x}^p$.
Human feedback is generated when 
a ranked gallery image $\bm{x}^g$ is selected 
by the user with a label 
$y \in \{\mbox{true}, \mbox{dissimilar}\}$ (Fig. \ref{fig:model_flow}(c)).
%
Once a feedback for probe $\bm{x}^p$ is received, 
the parameters of re-id model $f(\cdot)$ are updated
instantly (Fig. \ref{fig:model_flow}(d)) to 
re-order the gallery ranking list and give the user
immediate reward for the next feedback (Fig. \ref{fig:model_flow}(e)). 
{\bf (3)} When either a true match is found or a pre-determined maximum round of
feedback is reached,
the next probe is
presented for re-id matching in the gallery set. 
In contrast to pre-labelling training data required by
the conventional train-once-and-deploy {\em human-out-of-the-loop (HOL) re-id} scheme, {\em
	HIL re-id} has two unique characteristics: 
(a) Due to limited human
patience and labour budget~\cite{Hirzer_2011_SCIA},
a user typically prefers to examine only the top ranks rather than the
whole rank list,
and to provide only a few feedback.
(b) Instead of seeking to verify {\em positives} (true matches) for
each probe, which are most {\em unlikely} to appear in the
top ranks\footnote{In a large size gallery set, true matches are
	often scarce (only one-shot or few shots) and overwhelmed (appear in
	low-ranks) by false matches of high-ranks in the rank
	list.}, it is a much easier and more rewarding task for the user to
identify {\em strong-negatives}, that is, those top ranked negative
gallery instances {\em ``definitely not the one I am looking
  for''} -- visually very {\em dissimilar} to the target
image. 



Note that, in contrast to \cite{liu2013pop,wang2016human}, here we
consider a simpler human verification task by also ignoring {\em
  weak-negatives}: Those top ranked negative instances which {\em
  ``look similar but not the same person as I am looking for''}. The reasons are:
{\bf (1)} A user is inclined to notice strong negatives among the top ranks, 
i.e. a cognitively easier task (Fig.~\ref{fig:reid_feedbacks}(d)) due
to that most top ranks are likely to be weak negatives. Making correct
selection and verification of weak negatives requires much more effort. 
In contrast, a strong negative ``pops out'' readily to a user's
attention among the top ranks given the salience-driven visual
selective attention mechanism built into the human visual
system~\cite{parkhurst2002modeling}. 
{\bf (2)} We consider strong negatives in top ranks are {\em hard-unexpected negatives}:
``Hard'' since they are top-ranked negatives in the gallery thus
misclassified with high confidence (short matching
distance) to the wrong identity class by the current model;
``Unexpected'' since they are visually significantly dissimilar to the
probe image whilst among the top ranks, therefore violating
expectation and providing most informative feedback on model
mistakes\footnote{In this context, weak negatives in top ranks can be considered as {\em
  hard-expected negatives}~\cite{cui2016fine}.}.
Exploiting strong negatives to rectify model learning is more
cost-effective with less labelling required (Sec. \ref{sec:exp}).
Moreover, this is also compatible with the notion of salience-guided human eye movements
therefore more likely to encourage a user to engage with the re-id
task at hand whilst giving feedback, providing a higher degree of complementary effect 
between \change{machine incremental} learning from human feedback 
and human \change{instant} rewards from improved model output.


\subsection{Modelling Human Feedback as a Loss Function}
\label{sec:model_feedback}
\begin{figure}[!h]
	\centering
	\includegraphics[width=0.4\linewidth]{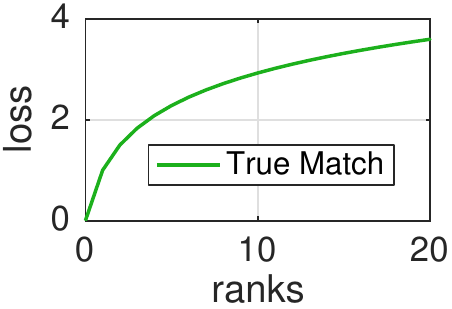}
	\hfil
	\includegraphics[width=0.4\linewidth]{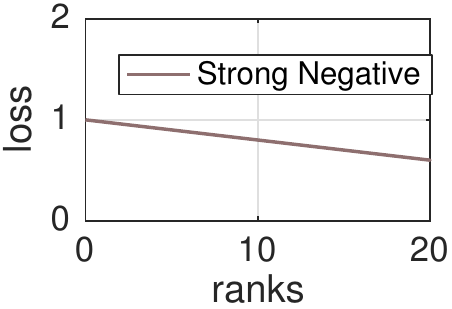}
	\vskip -0.3cm
	\caption{Values of $\mathcal{L}_y(k)$ for distinct human feedback, with $n_g = 50$. }
	\label{fig:loss}
	\vspace{-0.1cm}
\end{figure}

Formally, we wish to construct an incrementally optimised ranking function,
$f_{\bm{x}^p}(\bm{x}^g_i): \mathbb{R}^d \rightarrow \mathbb{R}$, 
where $f(\cdot)$ can be estimated by two types of human feedback
$y \in L = \{m, s\}$ as {\em true-match} and {\em
	strong-negative} 
respectively. Inspired by \cite{icml2014c2_lim14,WestonECML10,Usunier_2009_ICML},
we define a ranking error (loss) function for a feedback $y$ on a human
selected gallery sample $\bm{x}^g$ given a probe
$\bm{x}^p$ as:
\begin{equation}
err(f_{\bm{x}^p}(\bm{x}^g),y) = \mathcal{L}_y(rank(f_{\bm{x}^p}(\bm{x}^g))),
\label{err}
\end{equation}
where $rank(f_{x^p}(\bm{x}^g))$ denotes the rank of $\bm{x}^g$ given by $f_{\bm{x}^p}(\cdot)$, defined as:
\begin{equation}
rank(f_{\bm{x}^p}(\bm{x}^g)) = \sum_{\bm{x}^g_i \in \bm{G} \setminus \bm{x}^g}\mathcal{I}(f_{\bm{x}^p}(\bm{x}^g_i) \geqslant f_{\bm{x}^p}(\bm{x}^g)),
\end{equation}
where $\mathcal{I}(\cdot)$ is the indicator function. 
The loss function $\mathcal{L}_y(\cdot): \mathbb{Z}^+ \rightarrow \mathbb{R}^+$ 
transforms a rank into a loss. 
We introduce a novel re-id ranking loss defined as:
\begin{align}
&\mathcal{L}_y(k) = 
\begin{cases}
\sum_{i=1}^k \alpha_i, \;\;\;\; \;\text{if} \; y \in \{m\}\\
\sum_{i=k+1}^{n_g} \hat{\alpha}_i, \;  \text{if} \; y \in \{s\}
\end{cases}, \label{eqn:loss}
\\ &\text{with} \;\; \alpha_1 \geqslant \alpha_2 \geqslant \cdots \geqslant 0 , \;\; \text{and}
\;\; \hat{\alpha}_{n_g} \geqslant  \hat{\alpha}_{n_g-1} \geqslant \cdots \geqslant 0 . \nonumber
\end{align}
Note, different choices of $\alpha_i,\hat{\alpha}_i$ lead to distinct model responses to human feedback (Fig.~\ref{fig:loss}). 
We set $\alpha_i = \frac{1}{i}$ (large penalty with steep slope) when
$y$ indicates a {\em true-match} ($m$), 
and $\hat{\alpha}_i = \frac{1}{n_g - 1}$ with $n_g$ the gallery size (small penalty with gentle slope) 
when $y$ represents a {\em strong-negative} ($s$). 
Such a ranking loss is designed to favour a model update behaviour so that: (1)
{\em true-matches} are quickly pushed up to the top ranks, whilst (2)
{\em strong-negatives} 
are mildly moved towards the bottom
rank direction. Our experiments (Sec.~\ref{sec_eval:human}) show that
such a ranking loss criterion boosts very effectively the Rank-1
matching rate and pushes quickly {\em true-matches} to the top ranks at
each iteration of human feedback.

\subsection{\small Real-Time Model Update for Instant Feedback Reward}
\label{sec:model_update}
\noindent {\bf Model Formulation } Given the re-id ranking loss function defined in Eqn.~\eqref{eqn:loss},
we wish to have real-time model update to human feedback therefore
providing instant reward to user labour effort. 
To that end, we consider the HVIL re-id ranking model $f(\cdot)$ as a negative Mahalanobis distance metric:
\begin{equation}
f_{\bm{x}^p}(\bm{x}^g) = - \left[ (\bm{x}^p - \bm{x}^g)^\top \bm{M}
(\bm{x}^p - \bm{x}^g) \right], \; \bm{M}\in S^d_+ .
\label{eqn:matching}
\end{equation}
The positive semi-definite matrix $\bm{M}$ consists of model parameters to be learned
\change{from sequential human feedback}.

\vspace{0.1cm}
\noindent \textbf{Knowledge Cumulation by Online Learning }  
In the previous works~\cite{Hirzer_2011_SCIA,liu2013pop}, 
a re-id model $f(\cdot)$ is only optimised in isolation for each probe
without benefiting from previous feedback on other probes.
To overcome this limitation, we wish to optimise $f(\cdot)$
{\em incrementally} in an online manner~\cite{Chechik_2010_JMLR} 
for maximising the value of limited human feedback labour
budget. Moreover, to achieve real-time human-in-the-loop feedback and reward,
$f(\cdot)$ needs be 
estimated immediately on each human feedback.

Formally, given a new probe $\bm{x}^p_t$ at time step $t \in \{1,\cdots,\tau\}$
($\tau$ the pre-defined verification budget),
a user is presented with a gallery rank list computed by the previously estimated 
model $\bm{M}_{t-1}$ instead of a new ranking function re-initialised 
from scratch for this new probe. The user then verifies a
gallery image $\bm{x}^g_t$ in the top ranks with a label $y_t$,
generating a labelled triplet $(\bm{x}_t^p,\bm{x}_t^g, y_t)$.
Given Eqn.~\eqref{eqn:loss}, this triplet has a corresponding loss 
as $\mathcal{L}^{(t)} =
\mathcal{L}_{y_t}(rank(f_{\bm{x}^p_t}(\bm{x}^g_t)))$. We update the
ranking model by minimising the following object function: 
%
%
\begin{equation}
\bm{M}_{t} = \argmin_{\bm{M}\in S^d_+}\Delta_F(\bm{M},\bm{M}_{t-1}) + \eta  \mathcal{L}^{(t)},
\label{obj}
\end{equation}
where $\Delta_F$ is a Bregman divergence measure, 
defined by an arbitrary differentiable convex function $F$,
for quantifying the discrepancy between $\bm{M}$ and $\bm{M}_{t-1}$. 
The set $S^d_+$ defines a positive semi-definite (PSD) cone.
The tradeoff parameter $\eta\!\!>\!\!0$ balances the model update divergence and empirical ranking loss.
This optimisation updates the ranking model adopted from the
previous probe by encoding user feedback on the current probe.

\vspace{0.1cm}
\noindent \textbf{Loss Approximation for Real-Time Optimisation } 
In order to encourage and maintain user engagement in verification
feedback, real-time online incremental metric learning is
required. However, as $\mathcal{L}^{(t)}$ is discontinuous, the
overall objective function cannot be optimised efficiently by gradient-based learning methods.
We thus approximate the loss function by a continuous upper bound \cite{icml2014c2_lim14}
so that it is differentiable w.r.t. the parameter $\bm{M}$:
\begin{align}
\footnotesize
\widetilde{\mathcal{L}}^{(t)} = 
\frac{1}
{\mathcal{N}_t^-}
\sum_{\bm{x}^g_i \in \bm{G} \setminus \bm{x}_t^g} \mathcal{L}_{y_t}
\left( rank \left( f_{\bm{x}^p_t}(\bm{x}^g_t | \bm{M}_{t-1}) \right) \right) \nonumber \\
\cdot h_{y_t} \left( f_{\bm{x}^p_t}(\bm{x}^g_t | \bm{M}_{t}) -
f_{\bm{x}^p}(\bm{x}^g_i | \bm{M}_{t-1}) \right)^2 ,
\label{approx}
\end{align}
where $f_{\bm{x}^p_t}(\bm{x}^g_t|\bm{M}_{t-1})$ denotes the function value of $f_{\bm{x}^p_t}(\bm{x}^g_t)$ parametrised by $\bm{M}_{t-1}$, and
$h_{y_t}(\cdot)$ represents a hinge loss function defined as:
\begin{align}
\footnotesize
&h_{y_t}\big(f_{\bm{x}^p_t}(\bm{x}^g_t) - f_{\bm{x}^p_t}(\bm{x}^g_i)\big) = \nonumber \\
&\begin{cases}
max\big(0,1-f_{\bm{x}^p_t}(\bm{x}^g_t)+f_{\bm{x}^p_t}(\bm{x}^g_i)\big), \; \; \text{if} \; y_t \in \{m\}\\
max\big(0,1-f_{\bm{x}^p_t}(\bm{x}^g_i)+f_{\bm{x}^p_t}(\bm{x}^g_t)\big), \; \; \text{if} \; y_t \in \{s\}
\end{cases} .
\label{eqn:hinge}
\end{align}
The normaliser $\mathcal{N}_t^-$ in Eqn.~\eqref{approx} is the amount 
of violators, i.e. the gallery instances that generate non-zero hinge loss (Eqn.~\eqref{eqn:hinge})
w.r.t. the triplet $(\bm{x}_t^p,\bm{x}_t^g, y_t)$.


\vspace{0.1cm}
\noindent \textbf{Learning Speed-up by Most Violator Update } 
Given the loss approximation in Eqn.~\eqref{approx}, 
we can exploit the stochastic gradient descent (SGD) algorithm \cite{bottou_2010_large} 
for optimising the proposed model objective function Eqn.~\eqref{obj}
by iteratively updating on sub-sampled batches of all violators.
However, the computational overhead of iterative updates
can be high due to possibly large number of violators, 
and thus not meeting the real-time requirement. 
To address this problem, 
we explore a \textit{most violator update} strategy,
that is, to perform metric updates
using {\em only} the violator $\bm{x}^g_v$ with the most violation 
(Eqn.~\eqref{eqn:hinge}).
The final approximated empirical loss is then estimated as:
%
%
\begin{align}
\footnotesize
\widetilde{\mathcal{L}}^{(t)}_v = & \;
\mathcal{L}_{y_t} \left( rank \big(
f_{\bm{x}^p_t}(\bm{x}^g_t|\bm{M}_{t-1}) \big)
\right) \noindent \nonumber \\ 
& \cdot h_{y_t} \left( f_{\bm{x}^p_t}(\bm{x}^g_t | \bm{M}_t) -
f_{\bm{x}^p}(\bm{x}^g_v | \bm{M}_{t-1}) \right)^2.
\label{eqn:final_update}
\end{align}
Next, we derive $\bm{M}_t$ for updating the ranking metric.
Specifically, recall that the Bregman divergence 
between any two matrices $\bm{A}$ and $\bm{B}$ is defined as:
\begin{equation}
\Delta_F(\bm{A}, \bm{B}) = F(\bm{A}) - F(\bm{B}) - tr \big((\bm{A}-\bm{B})g(\bm{B})^\top\big),
\end{equation}
where $g(\cdot)$ denotes the derivative of $F$ (Eqn.~\eqref{obj})~\cite{tsuda2005matrix}
and $tr(\cdot)$ the matrix trace norm. 
After taking the gradient with the first argument $\bm{A}$, it has the following form:
\begin{equation}
\nabla_{\bm{A}} \Delta_F(\bm{A}, \bm{B}) = g(\bm{A}) - g(\bm{B}),
\end{equation}
By replacing $\mathcal{L}^{(t)}$ in Eqn.~\eqref{obj} with $\widetilde{\mathcal{L}}^{(t)}_v$, 
and setting the gradient of the minimisation objective in Eqn.~\eqref{obj} to zero, we have:
\begin{equation}
g(\bm{M}_t) - g(\bm{M}_{t-1}) + \eta
\nabla_{\bm{M}}\widetilde{\mathcal{L}}^{(t)}_v = 0.
\end{equation}
This gives the following ranking metric online updating criterion:
\begin{equation}
\bm{M}_{t} = g^{-1} \left( g(\bm{M}_{t-1}) - \eta
\nabla_{\bm{M}}\widetilde{\mathcal{L}}^{(t)}_v \right),
\label{update}
\end{equation}
where the gradient of $\widetilde{\mathcal{L}}^{(t)}_v$ w.r.t. $\bm{M}$ can be calculated as:
\begin{equation}
\nabla_{\bm{M}}\widetilde{\mathcal{L}}^{(t)}_v = \hat{\mathcal{L}}  (f_t - f_v - b_t) \bm{z}_t \bm{z}_t^\top,
\label{eqn:loss-gradient}
\end{equation}
with 
\begin{align}
\label{eq:notions}
\hat{\mathcal{L}} &= \mathcal{L}_{y_t} \left( rank \big(
f_{\bm{x}^p_t}(\bm{x}^g_t|\bm{M}_{t-1}) \big)
\right), \; 
f_v = f_{\bm{x}^p_t}(\bm{x}^g_v|\bm{M}_{t-1}),
 \\ 
f_t & = f_{\bm{x}^p_t}(\bm{x}^g_t|\bm{M}_{t}),  \;\; 
\bm{z}_t = \bm{x}_t^p - \bm{x}_t^g, \; 
b_t =\begin{cases}
1, & \text{if $y_t \in \{m\}$}. \\
-1, & \text{if $y_t \in \{s\}$}.
\end{cases}
\nonumber
\end{align}
For the convex function $F(\cdot)$, existing common choices include
squared Frobenius norm $\|\bm{M}\|_F^2$ and quantum entropy
$tr(\bm{M}\log(\bm{M}) - \bm{M})$.  The incremental update of the
HVIL model by Eqn.~\eqref{update} can be then optimised by a standard
gradient-based learning scheme such
as~\cite{icml2014c2_lim14,kivinen1997exponentiated,tsuda2005matrix}.  
%
%
In this work, we adopt a strictly convex function $F(\bm{M}) = -\log\det(\bm{M})$. 
This is because its gradient function $g(\cdot)$ is as simple as 
\begin{equation}
g(\bm{M}) = \nabla_{\bm{M}}F(\bm{M}) = \bm{M}^{-1},
\end{equation}
and along with Eqn. \eqref{eqn:loss-gradient} we can simplify Eqn. \eqref{update} as:
\begin{equation}
\bm{M}_{t} = \left( \bm{M}_{t-1}^{-1} - \eta
\hat{\mathcal{L}}  (f_t - f_v - b_t) \bm{z}_t \bm{z}_t^\top \right)^{-1}.
\end{equation}
Applying the Sherman Morrison formula~\cite{woodbury1950inverting},
we obtain the following online updating scheme for our HVIL model
$\bm{M}$:
\begin{equation}
\bm{M}_t = \bm{M}_{t-1} - \frac{\eta \hat{\mathcal{L}}  (f_t - f_v - b_t) \bm{M}_{t-1}\bm{z}_t \bm{z}_t^\top \bm{M}_{t-1} }
{1 + \eta \hat{\mathcal{L}}  (f_t - f_v - b_t) \bm{z}_t^\top \bm{M}_{t-1} \bm{z}_t}
\label{eqn:LEGO_update}
\end{equation}
%
To compute $\bm{M}_t$, we need to obtain the value of $f_t$
which however is parametrised by $\bm{M}_t$ (Eqn. \eqref{eq:notions}) and thus cannot be computed readily.
One potential optimisation option is resorting to gradient approximation
\cite{davis2007information}.
Instead, we propose to solve $\bm{M}_t$ with exact gradient for more accurate modelling, 
inspired by the LEGO metric update \cite{jain2009online}.
Specifically, by left multiplying $\bm{M}_t$ with $\bm{z}^\top$ and
right multiplying with $\bm{z}$, we obtain
\begin{equation}
\bm{z}^\top \bm{M}_t \bm{z} = f_t = \frac{\hat{f}}
{1 + \eta \hat{\mathcal{L}}  (f_t - f_v - b_t) \hat{f}}
\end{equation}
with $\hat{f} = f_{\bm{x}^p_t}(\bm{x}^g_t|\bm{M}_{t-1})$.
Then, $f_t$ can be solved by algebra transformation as:
\begin{equation}
f_t = \frac{\eta \hat{\mathcal{L}} (f_v+b_t) \hat{f} -1 + \sqrt{(\eta \hat{\mathcal{L}} (f_v+b_t) \hat{f} -1)^2 + 4 \eta \hat{\mathcal{L}} \hat{f}^2}  }{2 \eta \hat{\mathcal{L}}  \hat{f}}
\label{eqn:ft}
\end{equation}
Given this explicitly calculated $f_t $,
we can evaluate quantitatively Eqn.~\eqref{eqn:LEGO_update}
for online HVIL model updating. 
An overview of the HVIL online learning process is given in Algorithm~\ref{alg:HVIL}.
The updating scheme as described herein is favourable because it requires no 
computationally expensive eigen-decomposition to project the updated metric back to the
PSD cone, and the positive definiteness of $\bm{M}_t$ can be automatically guaranteed 
according to:
\vspace{0.1cm}
\noindent {\it \textbf{Theorem 1.} }
If $\bm{M}_{t-1}$ is positive definite, then $\bm{M}_{t}$ 
computed by Eqn. \eqref{eqn:LEGO_update} is also positive definite.

\vspace{0.1cm}
\noindent \textbf{\em Proof. } 
If $\bm{M}_{t-1}$ is a positive definite matrix, then 
\begin{equation}
\hat{f} = f_{\bm{x}^p_t}(\bm{x}^g_t|\bm{M}_{t-1}) = \bm{z}_t^\top \bm{M}_{t-1} \bm{z}_t > 0 \;\; \text{for all } \bm{z}_t. \nonumber
\end{equation}
Since $\eta > 0, \; \hat{\mathcal{L}} > 0$, we have 
\begin{equation}
\sqrt{(\eta \hat{\mathcal{L}} (f_v+b_t) \hat{f} -1)^2 + 4 \eta \hat{\mathcal{L}} \hat{f}^2} > |\eta \hat{\mathcal{L}} (f_v+b_t) \hat{f} -1|. \nonumber  
\end{equation}
Therefore, from Eqn. \eqref{eqn:ft} we have
\begin{equation}
f_t = f_{\bm{x}^p_t}(\bm{x}^g_t|\bm{M}_{t}) =  \bm{z}_t^\top \bm{M}_{t} \bm{z}_t > 0 \;\; \text{for all } \bm{z}_t. \nonumber
\end{equation}
Hence $\bm{M}_t$ is also a positive definite matrix.

\vspace{0.1cm}
\noindent {\bf Model Complexity }
This online HVIL model update by Eqn.~\eqref{eqn:LEGO_update} is
solved with a computational complexity of $\mathcal{O}(d^2)$ where $d$ is
the feature vector dimension, while
a cost of $\mathcal{O}(d^3)$ is required by most other schemes which
perform the Bregman projection back to the PSD cone~\cite{icml2014c2_lim14,kivinen1997exponentiated,tsuda2005matrix}.
%
Given all the components described above, our final model for Human
Verification Incremental Learning (HVIL) enables real-time incremental
model learning with human-in-the-loop feedback to model re-id rank
list. As shown in our evaluation (Sec.~\ref{sec_eval:human}), 
the proposed HVIL
model provides faster human-in-the-loop feedback-reward cycles 
as compared to alternative models.

\begin{algorithm}[h]
	\label{alg:HVIL}
	\footnotesize 
	\SetInd{0.1cm}{0.3cm}
	\caption{\footnotesize Human Verification Incremental Learning (HVIL) 
	}
	\KwIn{Unlabelled probe set $\mathcal{P}$ and gallery set $\mathcal{G}$;}
	\KwOut{Per probe optimised ranking lists; re-id models $\{\bm{M}_t\}_{t=1}^{\tau}$;}
	\vspace{0.1cm}
	{\bf Initialisation}: $\bm{M}_0$ = $\bm{I}$ (identity matrix, equivalent to the $L_2$ distance) \\
	\vspace{0.1cm}
	{\bf HIL person re-id:}\\
	\While{$ t < \tau$} 
	{
		Present the next probe $\bm{x}_t^p \in \mathcal{P}$; \\
		\tcp{maxIter: maximum interactions per probe}
		\For{$iter = 1:maxIter$}{ 
			Rank $\mathcal{G}$ with $\bm{M}_{t-1}$ against the probe $\bm{x}_t^p$ (Eqn.~\eqref{eqn:matching}); \\
			Request the human feedback $(\bm{x}_t^g, y_t)$; \\
			%
			Calculate $\widetilde{\mathcal{L}}^{(t)}_v$ with the most violator $\bm{x}^g_v$
			(Eqn.~\eqref{eqn:hinge} and \eqref{eqn:final_update}); \\
			$\bm{M}_t = update(\bm{M}_{t-1}, \widetilde{\mathcal{L}}^{(t)}_v)$ (Eqn.~\eqref{update}); \\
			%
		}
	}
	
	Return $\{\bm{M}_t\}_{t=1}^{\tau}$.
\end{algorithm}

%

\section{Metric Ensemble for HOL Re-Id}
\label{sec:method_ensemble}
Finally, we consider the situation when the limited human labour budget is exhausted at time $\tau$
and an automated HOL re-id strategy is required for any further probes as in conventional approaches.
In this setting, given that the HVIL re-id model is
optimised incrementally during the HIL re-id procedure, 
the latest model $\bm{M}_\tau$ optimised by the
human verified probe at time $\tau$ can be directly
deployed. However, it is desirable to construct an even ``stronger'' 
model based on metric ensemble learning. Specifically, a side-product
of HVIL is a series of models incrementally optimised {\em
	locally} for a set of probes with human feedback. We consider them as a set of
{\em globally} ``weak'' models $\{\bm{M}_j\}_{j=1}^{\tau}$, and wish to
construct a {\em single globally strong model} for re-identifying
further probes without human feedback.

\vspace{0.1cm}
\noindent {\bf Regularised Metric Ensemble Learning }
Given weak models $\{\bm{M}_j\}_{j=1}^\tau$, we compute a distance vector
$\bm{d}_{ij} \in \mathbb{R}^\tau$ for any probe-gallery pair ($\bm{x}^g_j$, ${\bm{x}_i^p}$):
\begin{equation} 
\bm{d}_{ij} = - \left[ f_{\bm{x}_i^p}(\bm{x}^g_j| \bm{M}_1), \cdots,
f_{\bm{x}_i^p}(\bm{x}^g_j| \bm{M}_\tau) \right]^\top
\end{equation}
The objective of metric ensemble learning is to obtain an optimal
combination of these distances for producing a single globally optimal
distance.
Here we consider the ensemble ranking 
function $f^{ens}_{\bm{x}^p_i}(\bm{x}^g_j)$ in a bi-linear form (shortened as $f^{ens}_{ij}$): 
%
%
\begin{equation}
\label{rankingscore}
f^{ens}_{ij} = f^{ens}_{\bm{x}^p_i}(\bm{x}^g_j) = - \bm{d}_{ij}^\top \bm{W} \bm{d}_{ij},
\quad \text{s.t.} \;\;\; \bm{W} \in S^\tau_{+} ,
\end{equation}
with $\bm{W}$ being the ensemble model parameter matrix that captures the correlations
among all the weak model metrics. In this context, previous work such as
\cite{Anton_2015_CoRR} is a special case of our model when $\bm{W}$ is
restricted to be diagonal only. 

\vspace{0.1cm}
\noindent \textbf{Objective Function } 
To estimate an optimal ensemble weights $\bm{W}$ with maximised identity-discriminative power, 
we re-use the true matching pairs verified during the human verification procedure 
(Sec.~\ref{sec:method_incre}) as ``training data'': $\mathcal{X}_{tr} = \{(\bm{x}^p_i, \bm{x}^g_i)\}_{i=1}^{N_l}$, 
and their corresponding person identities are denoted by $\mathcal{C}
= \{ c_i \}_{i=1}^{N_l}$. 
Note, ``training data'' here are only for
estimating the ensemble model weight, not for learning a distance metric. 
Since the ranking score $f^{ens}_{ij}$ in 
Eqn.~\eqref{rankingscore} is either negative or zero, 
we consider that in the extreme case, an {\em ideal} ensemble function $f^{*}_{ij}$ 
should provide the following ranking scores: 
\begin{equation}
f^{*}_{ij} =\begin{cases}
0, & \text{if $c_i = c_j$}, \\
-1, & \text{if $c_i \neq c_j$}.
\end{cases}
\end{equation}
Using  $\bm{F^*}$ to denote such an ideal ranking score matrix
and $\bm{F}^{ens}$ to denote an estimated score matrix by a 
given $\bm{W}$ with Eqn.~\eqref{rankingscore},
our proposed objective function for metric ensemble learning is then defined as:
%
\begin{equation}
\label{obj:ens}
\rho = \min_{\bm{W}} \|\bm{F}^{ens} - \bm{F^*}\|_{F}^2 + \nu \mathcal{R}(\bm{W}), 
\quad \text{s.t.} \;\;\; \bm{W} \in S^\tau_{+} ,
\end{equation} 
where $\|\cdot\|_F$ denotes a Frobenius norm, and
$\mathcal{R}(\bm{W})$ a regulariser on $\bm{W}$ with parameter $\nu$
controlling the regularisation strength.
Whilst common choices of $\mathcal{R}(\bm{W})$ include $L_1$, Frobenius norm, or matrix trace, 
we introduce the following regularisation for a Regularised Metric
Ensemble Learning (RMEL) re-id model:
\begin{equation}
\label{regu}
\mathcal{R}(\bm{W}) = - \sum_{i,j} f_{ij}^{ens}, \quad \; \text{if} \;\;\; c_i = c_j .
\end{equation}
Our intuition is to impose severe penalties for true match pairs with low ranking scores
since they deliver the most informative discriminative information for cross-view person re-id, whilst 
false match pairs are less informative.

\vspace{0.1cm}
\noindent \textbf{Optimisation } 
Eqn.~\eqref{obj:ens} is strictly convex with a guaranteed global
optimal so it can be optimised by any off-the-shelf toolboxes~\cite{cvx}.
We adopt the standard first-order projected gradient descent algorithm~\cite{Boyd_2004},
with the gradient of Eqn.~\eqref{obj:ens} computed as:
\begin{equation}
\nabla_{\bm{W}} = \sum_{i,j}(f^{*}_{ij} - f^{ens}_{ij}  + \nu\mathcal{I}[c_i = c_j])\bm{d}_{ij}\bm{d}_{ij}^\top,
\label{eqn:gradient_W}
\end{equation}
with $\mathcal{I}$ being the indicator function. 
Our optimisation algorithm is summarised in Algorithm \ref{alg:REML}.

\begin{algorithm}[h]
	\label{alg:REML}
	\footnotesize 
	\SetInd{0.1cm}{0.3cm}
	\caption{\footnotesize Regularised Metric Ensemble Learning (REML)}
	\KwIn{Training dataset $\mathcal{X}_{tr} = \{(\bm{x}^p_i, \bm{x}^g_i)\}_{i=1}^{N_l}$, 
		label set $\mathcal{C} = \{ c_i \}_{i=1}^{N_l}$, 
		learning rate $\epsilon$,
		max learning inteartion $\tau_\text{me}$,
		and weak HVIL models $\{\bm{M}_j\}_{j=1}^\tau$;}
	\KwOut{The optimal weight matrix $\bm{W}$ for the metric ensemble;}
	
	\vspace{0.1cm}
	{\bf Initialisation:} Randomly initialise $\bm{W}_0$ to some PSD matrix.  \\
	
	\vspace{0.1cm}
	{\bf Metric Ensemble Learning:}\\
	\For{$k = 1:\tau_\text{me}$}{ 
		Calculate gradient  $\nabla_{\bm{W}_{k-1}}$ (Eqn. \eqref{eqn:gradient_W}); \\
		Set $\bm{W}_{k} = \bm{W}_{k-1} - \epsilon \nabla_{\bm{W}_{k-1}}$; \\
		Perform eigen-decomposition of $\bm{W}_k$: $\bm{W}_{k} = \sum_{i} \lambda_i \bm{u}_i \bm{u}_i^\top$; \\
		Project $\bm{W}_k$ back to PSD cone:  \\
		$\bm{W}_k = \sum_i max(\lambda_i, 0)\bm{u}_i\bm{u}_i^\top$. \\
	}	
	Return $\bm{W}$.
\end{algorithm}

\vspace{0.1cm}
\noindent {\bf HOL Person Re-Id } 
Given the estimated optimal ensemble weight matrix $\bm{W}$ and the
weak models $\{\bm{M}_j\}_{j=1}^\tau$, a single strong ensemble model
(Eqn.~\eqref{rankingscore}) is made available for performing automated HOL
re-id of any further probes on the gallery population. Our experiments
(Sec.~\ref{sec:eval_auto_reid}) show that the proposed RMEL algorithm
achieves superior performance as compared to state-of-the-art 
supervised re-id models given the same amount of labelled data.

\section{Experiments}
\label{sec:exp}

Two sets of comparative experiments were conducted:
(1) The proposed HVIL model was evaluated under a {\em Human-In-the-Loop} (HIL) re-id setting
and an {\em enlarged} test gallery population was used to reflect real-world use-cases (Sec. \ref{sec_eval:human}). 
(2) In the event of limited human labour budget being exhausted and
human feedback becoming unavailable, the proposed HVIL-RMEL model was
evaluated under an automated {\em human-out-of-the-loop} (HOL) re-id setting
(Sec. \ref{sec:eval_auto_reid}).

\begin{figure} [!t] 
	\centering
	\subfigure[CUHK03]{
		\includegraphics[width=0.3\linewidth, height=0.3\linewidth]{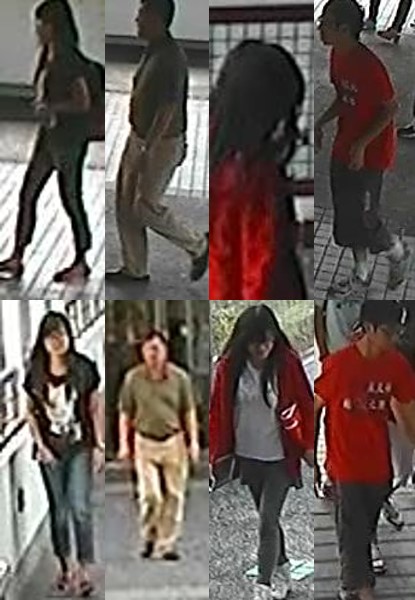}
	}
	\subfigure[Market-1501]{
		\includegraphics[width=0.3\linewidth, height=0.3\linewidth]{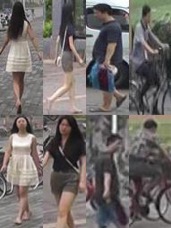}
	}
	\subfigure[VIPeR]{
		\includegraphics[width=0.3\linewidth, height=0.3\linewidth]{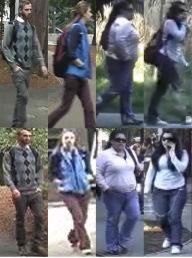}
	}
	\vskip -0.3cm
	\caption{Examples of cross-view person images from three person re-id datasets. 
		Two images in each column describe the same person.}
	\label{fig:dataset_img}
	\vspace{-0.1cm}
\end{figure}

\vspace{0.1cm}
\noindent {\bf Datasets } 
Two largest person re-id benchmarks:
CUHK03~\cite{Li_DeepReID_2014b} and
Market-1501~\cite{zheng2015scalable}, were chosen for evaluations due to
the need for large test gallery size.
CUHK03 contains 13,164 bounding box images of 1,360 people.
Two versions of person image are provided: manually labelled
and automatically detected, with the latter presenting more realistic 
detection misalignment challenges for practical deployments (Fig. \ref{fig:dataset_img}(a)).
We used both.
Market-1501 has 32,668 person bounding boxes of 1,501 people, obtained by automatic detection. 
Both datasets cover six outdoor surveillance
cameras with severely divergent and unknown viewpoints, illumination conditions, 
(self)-occlusion and background clutter (Fig. \ref{fig:dataset_img}(b)). 
In addition, we also selected the most common benchmark VIPeR~\cite{VIPeR} characterised 
with low imaging resolution and dramatic illumination variations (Fig. \ref{fig:dataset_img}(c)).
Compared to CUHK03 and Market-1501, VIPeR has a much smaller population size (632 people)
with fewer (1,264) labelled person images,
therefore only suitable for the conventional HOL re-id setting.
These three datasets present a wide range of re-id evaluation
challenges 
under different viewing conditions and with different population
sizes, as summarised in Table \ref{tab:dataset_stats}.

\begin{table}[h!] \footnotesize
	\centering
	\renewcommand{\arraystretch}{1}
	\setlength{\tabcolsep}{0.05cm}
	\vspace{-0.3cm}
	\caption{\footnotesize
		Settings of three person re-id datasets.
	}
	\vskip -.3cm
	\begin{tabular}{|c||c|c|c|c|c|c|}
		\hline
		Dataset  & 
		{Cams} &
		{IDs} & 
		{Labelled} & 
		{Detected} & HIL Split & HOL Split \\ \hline \hline %
		VIPeR \cite{VIPeR} & 2 & 632 & 1,264 & 0 & - & 316/316 \\  
		CUHK03 \cite{Li_DeepReID_2014b} & 6 & 1,467 & 13,164 & 13,164 & 1,000 & 360  \\
		Market-1501 \cite{zheng2015scalable} & 6 & 1,501 & 0 & 32,668 & 1,000 & 501 \\
		\hline
	\end{tabular}
	\label{tab:dataset_stats}
\end{table}

\vspace{0.1cm}
\noindent {\bf Data Partitions } 
For CUHK03 or Market-1501, we randomly selected 1,000 identities $D_{p1}$ ($p$ stands for population) 
as the partition to perform {\em HIL} re-id experiments.
The remaining partition of people $D_{p2}$ (360 on CUHK03, and 501 on Market-1501) were separated 
for evaluating the proposed model against state-of-the-art supervised re-id
methods for automated {\em HOL} re-id (see details in
Sec.~\ref{sec_eval:human} and Sec.~\ref{sec:eval_auto_reid}). Due to its small size, VIPeR was only used in the HOL experiments and the identities were split half-half for training and testing.
To obtain statistical reliability, we generated 10 different trials with different random partitions
  and reported their averaged results.

\vspace{0.1cm}
\noindent \textbf{Visual Features }
We adopted two types of image features:
{\bf (1)} The WHOS descriptor~\cite{KCCAReid}: 
A state-of-the-art {\em hand-designed} person re-id feature
(5,138 dimensions) composited by colour, HOG \cite{HOG_PersonDetect_CVPR12} 
and LBP \cite{LBP} histograms extracted from horizontal rectangular stripes\footnote{
The LOMO (26,960 dimensions) \cite{liao2015person} and GOG (27,622 dimensions) \cite{GOG} were not selected due to
their high dimensionality property which poses high computational cost for online model updating,
although they are possibly more discriminative.}.
{\bf (2)} The CNN feature learned by a recently proposed deep architecture for re-id \cite{geng2016deep}:
In contrast to hand-crafted WHOS features,
deep CNN features are extracted from a deep model trained by supervised learning from a large number of labelled training data.
Specifically, we trained the deep model with the entire person search
dataset \cite{xiao2016end}, which is independent of CUHK03, Market-1501
and VIPeR, therefore without any additional effect on their data
partitions. The trained deep model is directly deployed as a feature
extractor (1,024 dimensions) without any domain transfer learning by
fine-tuning on the three evaluation datasets.
Whilst adopting deep features from training a CNN model using labelled
data may seem to be inconsistent with the objective of this work --
eliminating the need for offline pre-collected training data, the main purposes of utilising the CNN feature are: 
(a) To evaluate the proposed HVIL on
different features; 
(b) To demonstrate any additional benefit of the proposed HVIL model
on a strong deep feature already learned from a large size labelled
training data.

\vspace{0.1cm}
\noindent {\bf Evaluation Metrics }
We adopted three performance evaluation metrics in the following experiments:
(1) Cumulative Match Characteristic (CMC): 
calculated as the cumulative recognition rate at each rank position. 
(2) Expected Rank (ER): 
defined as the average rank of all true matches.
(3) Mean Average Precision (mAP): 
first computing the area under the Precision-Recall curve for each probe, 
then calculating the mean of Average Precision over all probes.
For all HIL re-id models, we used the ranking result 
after the final human feedback applied on each probe. The averaged
results over all 10 trials were reported in comparisons. 

\subsection{Human-In-the-Loop Re-Id Evaluations}
\label{sec_eval:human}
%

\subsubsection{Experiment Settings}
\label{sec:human_protocol}
\noindent \textbf{Probe/Gallery Configuration }
For each of the $D_{p1}^i$ partitions,
we built a probe set for human operators to
perform HIL re-id.
In each trial, the probe set $\mathcal{P}^i$  
contains randomly selected
300 persons with one image per person.
For building the cross-view gallery set, we considered three different 
configurations to fully analyse the behaviour and scalability of the proposed HVIL method: 
{\bf (1)} {\em Single-shot gallery $\mathcal{G}^i_s$}: 
We randomly selected one cross-view image per person 
of all the 1,000 identities in partition $D_{p1}^i$ and construct
a single-shot gallery set $\mathcal{G}^i_s$ (1,000 person images) on both CUHK03 and Market-1501.
{\bf (2)} {\em Multi-shot gallery $\mathcal{G}^i_m$}:
We built the multi-shot gallery $\mathcal{G}^i_m$ by following \cite{zheng2015scalable}.
In particular, for all the 1,000 identities in partition $D_{p1}^i$, 
we used all cross-view images to construct the gallery set. 
As such, the average gallery image number is 4,919 on CUHK03 and 9,065 on Market-1501.
Note that, we did not utilise the label information about which gallery images are of the same person, 
and thus both CMC and mAP can be used for performance evaluation.
{\bf (3)} {\em Open-world gallery $\mathcal{G}^i_d$}:
We considered a more challenging 
setting with a large number of distractors involved in the gallery set. 
Specifically, we added 34,574
bounding boxes of 11,934 persons from the person search dataset~\cite{xiao2016end} to 
the single-shot gallery set $\mathcal{G}^i_s$. 
The resulted gallery $\mathcal{G}^i_d$ size is 35,574 on both datasets. 
This is to evaluate the scalability of HIL re-id methods when operating under the open-world re-id setting
featured with a huge gallery search space.

\vspace{0.1cm}
\noindent \textbf{Human Feedback Protocol }  
Human feedback were collected on all 10 trials 
of $D_{p1}^i$ partitions and all 3 different gallery configurations, in total
$3\times10=30$ independent 
sessions on each dataset by 5 volunteers as users.
During each session, a user was asked to 
perform the {\em HIL} re-id
on probes in the probe set $\mathcal{P}^i$ 
against a gallery set $\mathcal{G}^i \in \{\mathcal{G}^i_s,\mathcal{G}^i_m,\mathcal{G}^i_d\}$.
For each probe person, a \textit{maximum} of 3 rounds of user interactions are allowed. 
We limited the users to verify only the top-50 in the rank list ($5\%$ of $\mathcal{G}^i_s$, $0.5\sim1\%$ of $\mathcal{G}^i_m$, and $0.1\%$ of $\mathcal{G}^i_d$).
During each interaction: (1) A user selects one gallery image as
either \textit{strong-negative} or \textit{true-match}; and (2) the 
system takes the feedback, updates the ranking function 
and returns the re-ordered ranking list, all in real-time (Sec.~\ref{sec:method_incre}). 
The HVIL model was evaluated against \change{ten} existing models for
	HIL re-id deployment as follows.

\begin{table*} [!t] 
	\centering
	\caption{
		Human-in-the-loop person re-id with {\bf single-shot}
                galleries. Gallery Size: 1,000 for both CUHK03 and
                Market-1501. {\bf L}: Labelled; {\bf D}: Detected.
	}
	\label{tab:HIL_single}
	\vskip -0.3cm
	\scalebox{.95}{
		\renewcommand{\arraystretch}{1}
		\setlength{\tabcolsep}{0.2cm}
		\begin{tabular}{|l|ccc|ccc|ccc|ccc|ccc|ccc|}
			\hline
			Feature 
			& \multicolumn{9}{c|}{WHOS~\cite{KCCAReid} }
			& \multicolumn{9}{c|}{CNN~\cite{geng2016deep}
                          (except for DGD and Inception-V3)}
		    \\
			\hline
			Dataset 
			& \multicolumn{3}{c|}{CUHK03 (L)} 
			& \multicolumn{3}{c|}{CUHK03 (D)} 
			& \multicolumn{3}{c|}{Market-1501 (D)}
			& \multicolumn{3}{c|}{CUHK03 (L)} 
			& \multicolumn{3}{c|}{CUHK03 (D)} 
			& \multicolumn{3}{c|}{Market-1501 (D)}
            \\
			\hline
			Rank (\%)   
			& 1         & 50         & 100      
			& 1         & 50         & 100     
			& 1         & 50         & 100      
			& 1         & 50         & 100      
			& 1         & 50         & 100      
			& 1         & 50         & 100     
			\\ \hline \hline
			L2      
			& 2.9       & 31.1      & 43.2  
			& 2.7 & 29.8 & 41.6        
			&  16.1     & 66.6       &  76.6      
			& 19.0 & 72.0 & 82.3
			& 17.1 & 67.0 & 78.1
			& 44.2 & 94.4 & 97.5 	
			\\ 
			kLFDA~\cite{xiong2014person}   
			& 5.9       & 47.3      & 60.1    
			& 4.7 & 39.6 & 51.7  
			&  21.8     & 85.8       & 91.5    
			& 21.4 & 77.4 & 86.2 
			& 19.4 & 73.7 & 82.7
			& 52.9 & \bf  97.2 &  \bf 98.5 
			\\ 
			XQDA \cite{liao2015person}   
			& 3.7       & 40.2      & 53.6    
			& 2.4 & 22.4 & 33.3
			&  18.3     & 75.1       &  83.5     
			& 19.8 & 76.9 & 85.8 
			& 17.7 & 73.9 & 83.0
			&49.6& 97.0 & \bf  98.5 
			 \\ 
			MLAPG \cite{Liao_2015_ICCV}  
			& 4.2       & 39.5      & 52.4     
			& 3.5 & 36.1 & 49.3
			&  24.1     & 84.5       &  91.2     
			&11.8 & 69.6 & 82.5 
			& 10.2 & 64.3 & 77.9
			& 37.7 & 95.5 & 97.9 
			\\
			NFST \cite{zhang2016learning}
			& 7.1 & 41.5 & 54.7
			& 4.9 & 37.4 & 48.5
			& 34.4 & 85.3 & 90.7
			& 9.9 & 41.7 & 51.3
			& 9.5 & 38.0 & 47.8
			& 45.0 & 89.7 & 93.3
			\\
			HER \cite{wang2016highly}
			& 7.6 & 46.0 & 58.1
			& 5.7 & 41.8 & 53.8
			& 39.1 & \bf 90.8 & \bf94.7
			& 16.2 & 73.5 & 84.3
			& 14.5 & 69.9 & 80.2
			& 44.0 & 96.1 & 98.3\\ 
            DGD \cite{xiao2016learning}
            & - & - & -
            & - & - & -
            & - & - & -
            & 12.0 & 58.0 & 69.8 
            & 10.1 & 49.8 & 61.6
            & 58.4 & 95.7 & 97.4
            \\
            Inception-V3 \cite{szegedy2016rethinking}
            & - & - & -
            & - & - & -
            & - & - & -
            & 15.7 & 63.7 & 74.4
            & 15.3 & 62.5 & 72.2
            & 51.6 & 94.7 & 96.8
            \\			
			\hline
			EMR~\cite{xu2011efficient}     
			& 29.3 & 29.3 & 40.7 
			& 27.7 & 27.7 & 39.5
			& 64.2 & 64.2 & 74.2
			& 73.5 & 73.5 & 83.7
			& 66.7 & 66.7 & 77.5
			& 92.7 & 92.7 & 96.8
			\\ 
			Rocchio~\cite{lin2015effect} 
			& 32.0 & 38.7 & 46.2 
			& 29.0 & 36.2 & 43.8
			& 61.7 & 70.2 & 77.5
			& 62.0 & 79.2 & 85.2 
			& 56.2 & 74.3 & 80.8 
			& 81.2 & 94.5 & 93.3 
			\\ 
			POP \cite{liu2013pop}    
			& 44.0 & 51.5 & 60.0 
			& 41.7 & 48.5 & 58.8
			& 75.0 & 78.5 & 84.5
			& 74.7 & 74.8 & 77.2
			& 69.0 & 70.7 & 73.2 
			& 92.8 & 93.0 & 93.3 
			\\ \hline \hline
			\bf HVIL (Ours)     
			&\bf 60.2 & \bf 68.2 & \bf 78.5 
			& \bf 53.7 & \bf 65.0 & \bf 75.3
			& \bf 84.5 & 89.2 & 93.2 
			&\bf 84.2 & \bf 89.2 & \bf 93.3 
			& \bf 80.3 & \bf 86.0 & \bf 91.2
			& \bf 95.3 & 96.0 & 98.3 
			\\ \hline
	\end{tabular}}
\vspace{-0.1cm}
\end{table*}

\begin{table*} [!t] 
	\centering
	\caption{
		Human-in-the-loop person re-id with {\bf multi-shot}
                galleries. Gallery Size: 4,919 for CUHK03 and 9,065
                for Market-1501.
                {\bf L}: Labelled; {\bf D}: Detected.
	}
	\label{tab:HIL_multi}
	\vskip -0.3cm
	\scalebox{1}{
		\renewcommand{\arraystretch}{1}
		\setlength{\tabcolsep}{0.35cm}
		\begin{tabular}{|l|cc|cc|cc|cc|cc|cc|}
			\hline
			Feature 
			& \multicolumn{6}{c|}{WHOS~\cite{KCCAReid} }
			& \multicolumn{6}{c|}{CNN~\cite{geng2016deep}
                          (except for DGD and Inception-V3) }  \\
			\hline
			Dataset 
			& \multicolumn{2}{c|}{CUHK03 (L)} 
			& \multicolumn{2}{c|}{CUHK03 (D)} 
			& \multicolumn{2}{c|}{Market-1501 (D)} 
			& \multicolumn{2}{c|}{CUHK03 (L)} 
			& \multicolumn{2}{c|}{CUHK03 (D)} 
			& \multicolumn{2}{c|}{Market-1501 (D)} \\
			\hline
			Rank (\%)   
			& R-1     &  mAP    & R-1     & mAP    
			& R-1     &  mAP    & R-1     & mAP   
			& R-1     &  mAP    & R-1     & mAP       
			\\ \hline \hline
			L2      
			& 4.1 & 14.1 
			& 3.6 & 13.9 
			&  28.0 & 23.9  
			& 22.0 & 29.5 
			& 20.7 & 28.0 
			& 58.0 & 50.9     
			\\ 
			kLFDA~\cite{xiong2014person}   
			& 8.1 & 17.8 
			& 6.3 & 16.5 
			& 47.1 & 39.9 
			& 25.4 & 32.8 
			& 23.9 & 31.0 
			& 67.7 & 63.0    
			\\ 
			XQDA \cite{liao2015person}   
			& 3.6 & 14.9 
			& 4.5 & 14.5 
			& 34.3 & 30.1  
			& 24.5 & 31.7 
			& 22.5 & 30.0 
			& 63.4 & 58.1
			 \\ 
			MLAPG \cite{Liao_2015_ICCV}  
			& 5.0 & 15.1 
			& 5.1 & 15.1 
			& 44.3 & 40.8  
			& 14.8 & 23.7
			& 12.2 & 21.9 
			& 54.5 & 50.8
			\\
			NFST \cite{zhang2016learning}
			& 8.2 & 17.5
			& 7.7 & 16.6
			& 68.3 & 62.1
			& 20.2 & 26.8
			& 18.6 & 25.3
			& 76.2 & 69.9
			\\			
			HER \cite{wang2016highly}
			& 9.5 & 18.6
			& 8.1 & 17.4
			& 68.9 & 61.7
			& 24.3 & 31.4
			& 22.3 & 29.3
			& 77.4 & 72.1\\ 
            DGD \cite{xiao2016learning}
            & - & - 
            & - & -
            & - & -
            & 15.1 & 23.5
            & 13.0 & 21.4
            & 82.1 & 75.9
            \\
            Inception-V3 \cite{szegedy2016rethinking}
            & - & - 
            & - & -
            & - & -
            & 19.2 & 27.1
            & 18.3 & 26.2
            & 76.3 & 71.4
            \\
			\hline
			EMR~\cite{xu2011efficient}         
			& 30.8 & 20.2
			& 29.7 & 19.3
			& 76.0 & 31.7 
			& 71.3 & 40.6  
			& 66.3 & 37.5 
			& 94.0 & 57.7
			 \\ 
			Rocchio~\cite{lin2015effect} 
			& 34.0 & 26.4 
			& 30.7 & 23.7 
			& 74.3 & 37.1  
			& 59.3 & 50.0  
			& 56.0 & 46.8 
			& 83.7 & 65.1
			       \\ 
			POP \cite{liu2013pop}    
			& 43.0 & 39.4  
			& 44.3 & 38.2 
			& 82.7 & 52.7
			& 71.7 & 68.2  
			& 68.0 & 64.3 
			& 94.0 & 74.0    
			\\ \hline \hline
			\bf HVIL (Ours)     
			& \bf 63.0 & \bf 59.0 
			& \bf 53.7 & \bf 48.7 
			& \bf 87.3 & \bf 63.3 
			& \bf 84.0 & \bf 73.4 
			& \bf 80.7 & \bf 72.7 
			& \bf 96.0 & \bf 83.3      
			\\ \hline
	\end{tabular}}
\vspace{-0.1cm}
\end{table*}

\vspace{0.1cm}
\noindent {\bf HIL Competitors } 
Three existing HIL models were compared: 
(1) POP~\cite{liu2013pop}: 
The current state-of-the-art HIL re-id
method based on Laplacian SVMs and graph label propagation;
(2) Rocchio~\cite{lin2015effect}: 
A probe vector modification model updates iteratively the probe's feature vector based on 
human feedback, widely used for image retrieval tasks~\cite{datta2008image};
(3) EMR~\cite{xu2011efficient}: 
A graph-based ranking model that optimises the ranking function by least square regression. 
For a fair comparison of all four HIL
models, the users were asked to verify the same probe and gallery data
$(\mathcal{P}^i, \mathcal{G}^i)$ with the same two types of feedback given
the ranking-list generated by each model.

\vspace{0.1cm}
\noindent {\bf HOL Competitors } 
In addition, seven state-of-the-art conventional HOL supervised
learning models were also compared:
kLFDA~\cite{xiong2014person}, 
XQDA \cite{liao2015person}, 
MLAPG \cite{Liao_2015_ICCV}, 
NFST \cite{zhang2016learning},
HER \cite{wang2016highly},
DGD \cite{xiao2016learning},
and Inception-V3 \cite{szegedy2016rethinking}, among them two are deep
learning models (DGD and Inception-V3). 
These supervised re-id methods were trained using fully pre-labelled data in the separate partition  $D_{p2}^i$ 
(CUHK03: averagely 3,483 images of 360 identities; Market-1501: averagely 7,737 images of 501 identities)
before being deployed to 
$\mathcal{P}^i$ and $\mathcal{G}^i$ for automated HOL re-id testing. 
Note, the underlying human labour
effort for pre-labelling the training data to learn these
supervised models was significantly greater -- exhaustively searching
3,483 and 7,737  {\em true} matched images
respectively for CUHK03 and Market-1501, than that required by the
HIL methods -- 
between 300 to 900 {\em indicative} verification (strong negative
or true match) given a maximum of 300 probes 
\change{with each allocated a maximum of 3 feedback}
on both CUHK03 and Market-1501, so only 1/10th of and weaker user input
than supervised HOL models. 
It should be noted that non-deep distance metric models (kLFDA, XQDA,
MLAPG, NFST, HER) were trained using either hand-crafted
WHOS~\cite{KCCAReid} or deep learning CNN~\cite{geng2016deep}
features (Sec.~\ref{sec:exp}), while DGD and Inception-V3 were trained
directly from raw images in $D_{p2}^i$ 
\change{with the intrinsic capability of learning}
their own deep CNN features (256 dimensions for DGD and 2,048
for Inception-V3).

\vspace{0.1cm}
\noindent {\bf Implementation Details } 
For implementing the HVIL model (Sec.~\ref{sec:method_incre}), 
the only hyper-parameter $\eta$ (Eqn.~\eqref{obj}) was set to 0.5 on both CUHK03 and Market-1501.
We found that HVIL 
is insensitive to $\eta$ with a wide satisfiable range from $10^{-1}$ to $10^1$. 
For POP, EMR, and Rocchio, we adopted 
the authors' recommended parameter settings as in \cite{liu2013pop,xu2011efficient,lin2015effect}.
For all HIL methods above, we applied $L_2$ distance as the initial ranking function
$f_0(\cdot)$ without loss of generalisation\footnote{
	No limitation on considering any other distance or similarity metrics, either learned or not.
	However, non-learning based generic metrics are more scalable and transferable in real-world.
}. 
Note that for  HVIL, once $f_0(\cdot)$ was
initialised for only the very first probe, 
it was then optimised incrementally across different probes. 
In contrast, for POP and EMR and Rocchio,
each probe had its own $f_0(\cdot)$ initialised as $L_2$ 
since the models are not cumulative across different probes.
For HOL competitors, the parameters 
were determined by cross-validation on $D_{p2}$ with the authors'
published codes. All the models except DGD and Inception-V3 used the same
two feature descriptors for 
comparison (WHOS \cite{KCCAReid} and CNN feature
\cite{geng2016deep}). DGD~\cite{xiao2016learning} and
Inception-V3~\cite{szegedy2016rethinking} used their own
deep features from training their CNN networks.


\subsubsection{Evaluations on Person Re-Id Performance}

\begin{table*} [!t] 
\centering
\caption{
	Human-in-the-loop person re-id with {\bf open-world} galleries
        consisting of 34,574 {\bf distractors}. Gallery Size: 35,574
        for both CUHK03 and Market-1501. 
        {\bf L}: Labelled; {\bf D}: Detected. 
        Note: POP results are unavailable because it was
        {\em intractable} on our computing hardware.  
}
\vskip -0.3cm
\label{tab:HIL_distractor}
\scalebox{.95}{
	\renewcommand{\arraystretch}{1.03}
	\setlength{\tabcolsep}{0.2cm}
	\begin{tabular}{|l|ccc|ccc|ccc|ccc|ccc|ccc|}
	\hline
	    Feature & \multicolumn{9}{c|}{WHOS~\cite{KCCAReid}} &
            \multicolumn{9}{c|}{CNN~\cite{geng2016deep} (except for
              DGD and Inception-V3)} \\
		\hline
		Dataset 
		& \multicolumn{3}{c|}{CUHK03 (L) } 
		& \multicolumn{3}{c|}{CUHK03 (D) } 
		& \multicolumn{3}{c|}{Market-1501 (D)} 
		& \multicolumn{3}{c|}{CUHK03 (L)} 
		& \multicolumn{3}{c|}{CUHK03 (D) } 
		& \multicolumn{3}{c|}{Market-1501 (D)}\\
		 \hline
		Rank (\%)   
		& 1         & 50        & 100        
		& 1         & 50         & 100       
		& 1         & 50        & 100      
		& 1         & 50         & 100    
		& 1         & 50        & 100      
		& 1         & 50         & 100    
		\\ \hline \hline
		L2      
		& 2.8 & 27.2 & 38.2 
		& 2.6 & 24.8 & 34.4
		& 10.7 & 43.9 & 51.5 
		& 18.3 & 69.6 & 80.1 
		& 16.6 & 65.0 & 75.9
	    & 31.4 & 77.6 & 84.0   \\ 
		kLFDA~\cite{xiong2014person}   
		& 5.6 & 32.9 & 44.8 
		& 3.6 & 28.1 & 38.0 
        & 19.8 & 67.6 & 76.1
        & 17.7 & 66.9 & 77.1
        & 16.8 & 63.6 & 72.9	
        & 38.4 & 84.2 & 89.7
		\\ 
		XQDA \cite{liao2015person}   
		& 3.1 & 25.3 & 36.3 
		& 2.4 & 21.7 & 32.0
		& 16.6 & 61.9 & 70.7 
		& 15.5 & 61.7 & 70.6
		& 13.2 & 58.1 & 67.7  
		& 31.3 & 77.0 & 84.4	  \\ 
		MLAPG \cite{Liao_2015_ICCV}  
		& 3.7 & 33.3 & 44.0 
		& 2.8 & 28.9 & 39.2
		& 18.9 & 67.6 & 76.3 
		& 6.4 & 34.6 & 43.1  
		& 5.8 & 30.2 & 37.9   
		& 20.1 & 65.0 & 74.0   \\ 
		NFST \cite{zhang2016learning}
		& 5.6 & 34.6 & 45.6
		& 4.2 & 30.3 & 40.5
		& 30.1 & 78.3 & 85.0
		& 9.8 & 41.4 & 51.0	
		& 9.4 & 37.8 & 47.5
		& 39.6 & 83.7 & 88.4\\		
		HER \cite{wang2016highly}
		& 6.3 & 36.2 & 46.0
		& 4.5 & 31.4 & 40.5
		& 32.7 & 80.8 & 86.0
		& 12.3 & 57.3 & 66.5
		& 11.8 & 54.7 & 64.1
		& 26.1 & 70.6 & 79.0	
		\\ 
        DGD \cite{xiao2016learning}
            & - & - & -
            & - & - & -
            & - & - & -
            & 7.2 & 29.1 & 35.0
            & 5.6 & 23.2 & 29.0
            & 48.6 & 86.3 & 89.2
        \\		
        Inception-V3 \cite{szegedy2016rethinking}
        & - & - & -
        & - & - & -
        & - & - & -
        & 8.9 & 31.5 & 38.2
        & 7.4 & 30.5 & 37.6
        & 37.0 & 79.4 & 83.9
        \\
		\hline
		EMR~\cite{xu2011efficient}     
		& 25.8 & 25.8 & 35.5
		& 23.1 & 23.1 & 32.2
		& 40.8 & 40.8 & 46.8   
		& 70.7 & 70.7 & 81.0 
		& 66.3 & 66.3 & 77.7
		& 72.7 & 72.7 & 80.7 \\ 
		Rocchio~\cite{lin2015effect} 
		& 28.7 & 32.3 & 37.5 
		& 25.3 & 30.0 & 37.0
		& 43.6 & 46.2 & 48.8
		& 61.0 & 74.0 & 81.7 
		& 56.7 & 73.7 & 80.0
		& 64.3 & 74.3 & 80.0   
		\\ 
        POP~\cite{liu2013pop}
        & - & - & - 
		& - & - & -
		& - & - & -
		& - & - & -
		& - & - & -
		& - & - & -
        \\ \hline \hline
		\bf HVIL (Ours)     
		& \bf 55.6 & \bf 65.7 & \bf 74.8 
		& \bf 52.0 & \bf 60.3 & \bf 67.8
		& \bf 61.7 & \bf 70.8 & \bf 76.7
		& \bf 80.3 & \bf 86.0 & \bf 91.3 
		& \bf 73.3 & \bf 84.7 & \bf 89.3 
		& \bf 91.3 & \bf 93.3 & \bf 96.0 \\ \hline
	\end{tabular}}
\vspace{-0.1cm}
\end{table*}

The person re-id performances of all HIL and HOL methods on
$\mathcal{P}^i$ and $\{\mathcal{G}^i_s,\mathcal{G}^i_m,\mathcal{G}^i_d\}$ 
are compared in 
Tables \ref{tab:HIL_single} (single-shot), 
\ref{tab:HIL_multi} (multi-shot), and   
\ref{tab:HIL_distractor} (open-world) respectively.  

\vspace{0.1cm}
\noindent \textbf{HIL vs. HOL Re-Id Methods }
We first compared the re-id matching performance of HIL and HOL re-id schemes.
It is evident from the three Tables that
the HIL methods outperform significantly the conventional HOL counterparts
in all testing settings on both datasets.
Specifically, in single-shot setting (Table \ref{tab:HIL_single}),
{\em all} conventional supervised re-id models suffered severely
when the gallery size was enlarged to 1,000
from their standard setting.
For example,
the state-of-the-art deep re-id model DGD~\cite{xiao2016learning} 
can achieve $72.6\%$ Rank-1 rate on CUHK03 (Labelled) under
the test protocol of using the 100-sized test gallery. 
However, its Rank-1 accuracy drops dramatically to only $12.0\%$
Rank-1 on CUHK03 (Labelled) and $10.1\%$ (Detected) under the
1,000-sized test gallery evaluated here.
Similar performance drops occur for all other HOL models. 
Such low Rank-1 matching accuracies show that, existing best supervised re-id approaches 
are still far from being sufficiently mature to provide a fully automated HOL re-id solution in real world. 
On the contrary,
HIL methods make more realistic assumptions by considering human in the loop, 
and leverage limited human efforts to directly drive up model matching performance
by mining the joint human-machine benefits. 
The advantage in re-id matching by the HOL methods is clear:
for example, with WHOS feature the proposed HVIL achieves over $50\%$ and $80\%$ in Rank-1 on CUHK03 and Market-1501 (Table \ref{tab:HIL_single}), which is much more acceptable in practical use.
In terms of supervision cost, the supervised HOL models were offline trained 
on a large-sized pre-labelled data in $D_{p2}$
with an average of 3,483 cross-view images of 360 identities on CUHK03, 
and 7,737 images of 501 identities on Market-1501.
Whereas the HIL models required much less human verification effort, 
e.g. at most 3 feedback for each probe in top-50 ranks only, in total $300\sim900$ weak feedback. 
Human feedback is neither restricted to be only true matches, 
nor exhaustively labelling person identity labels, 
nor searching true matches in a huge image pool. 
These evidences suggest that HIL re-id
is a more cost-effective and promising scheme in 
exploiting human effort for real-world applications
as compared to the conventional HOL re-id approach.

Among all HIL re-id models, the proposed HVIL 
achieves the best performance. For instance, it is found in Table \ref{tab:HIL_single} that 
the HVIL improves significantly over the state-of-the-art HIL model POP on Rank-1 score, e.g.
from $44.0\%$ to $60.2\%$ on CUHK03 (Labelled), 
from $41.7\%$ to $53.7\%$ on CUHK03 (Detected), 
and from $75.0\%$ to $84.5\%$ on Market-1501, when the WHOS feature is used.
HVIL's advantage continues over all ranks. 
This demonstrates the compelling advantages of the HVIL model in cumulatively exploiting human verification feedback, whilst other existing human-in-the-loop models have no mechanisms for 
sharing human feedback knowledge among different probes.

\vspace{0.1cm}
\noindent \textbf{Effect of Features }
Next, we evaluated the effect of different visual features by comparing the hand-crafted WHOS~\cite{KCCAReid} and the most recent deep CNN feature~\cite{geng2016deep} learned from the large scale person search dataset~\cite{xiao2016end}.
As shown in Table \ref{tab:HIL_single}, 
the CNN feature is much more discriminative and view-invariant than the WHOS
thanks to the access of large quantity of labelled data and the strong deep \change{representation} learning capacity.
Specifically, with CNN feature, even the generic L2 metric can achieve $19.0\%$/$17.1\%$ and $44.2\%$ on CUHK03 (Labelled/Detected) and Market-1501, respectively.
Importantly, CNN feature can be well complementary with HIL re-id methods: 
The HIL re-id Rank-1 rates are further boosted to a more satisfying level, e.g. $84.2\%$/$80.3\%$ and $95.3\%$ 
by the proposed HVIL. 
This implies the great compatibility of the HVIL with deep feature learning.
%
%
On the other hand, it is found that with such a powerful deep CNN feature, 
HOL models are still outperformed drastically by HIL methods.
This suggests the consistent and general advantages of the HIL re-id scheme 
over the HOL approach given various types of visual features.

\vspace{0.1cm}
\noindent {\bf Single-Shot vs. Multi-Shot }
We evaluated the effect of shot number in the gallery set in person re-id performance.
When more shots of a person are available (Table \ref{tab:HIL_multi} vs. Table \ref{tab:HIL_single}),
re-id matching accuracy can be improved in most cases by either HIL and HOL methods including the proposed HVIL.
However, the best results are still generated by the HVIL model.
This suggests the steady advantage of the proposed method in different search gallery settings.
In particular, we have the following observations and justifications:
(1) The Rank-1 improvement degree varies over different datasets,
with Market-1501 benefiting more than CUHK03.
The plausible reason is that, Market-1501 person images 
give more pose and detection misalignment challenge due to poorer person bounding box detection,
and therefore multi-shot images with various poses and detection qualities can bring more gains.
(2) The HVIL model seem to benefit less from multi-shot gallery images as compared to other methods.
This may be due to the better capability of mitigating the pose/detection misalignment challenge by the proposed incremental model learning, thus not needing multiple shots as much as the other models do.

\vspace{0.1cm}
\noindent \textbf{Effect of Distractors in Open-World Setting } Finally, we
evaluated the effect of open-world distractors in the gallery set 
for further testing the model scalability.
This evaluation is made by comparing Table \ref{tab:HIL_single} and Table \ref{tab:HIL_distractor}.
After adding 34,574 person bounding boxes as distractors to the 1000 sized single-shot gallery 
(i.e. the gallery size is enlarged by $35$ times), 
we observed that (1) As expected, all methods suffered from some drop in re-id performance;
(2) The HIL methods outperform more significantly the HOL models under the open-world setting;
and (3) the proposed HVIL again achieves the best re-id performance, and particularly on the CUHK03 (Detected) dataset, the addition of 34K distractors causes only a $1.7\% = 53.7\%-52.0\%$ Rank-1 drop.
This again suggests the clear advantages and superiority of having human in the loop for real-world person re-id applications when the gallery population size is inevitably large in the open-world operation scenarios. 
More specifically, when the WHOS feature was used, 
the best HOL model HER's Rank-1 rates dropped from $7.6\%$ to $6.3\%$,
$5.7\%$ to $4.5\%$, and $39.1\%$ to $32.7\%$ on CUHK03 (Labelled), CUHK03 (Detected), and Market-1501 respectively. 
The best HIL competitor, POP, completely fails to operate with such a large gallery set. 
The reason is that POP requires to build an affinity graph and calculate the graph Laplacian on all the gallery samples to propagate human labels. 
Given a 34,574-sized gallery set, the affinity graph alone takes
$4.78$ GB storage which is both difficult to process (out of memory) for
common workstations and suffering from slow label propagation.
\begin{figure} [!t] 
	\centering
	\includegraphics[width=0.156\textwidth]{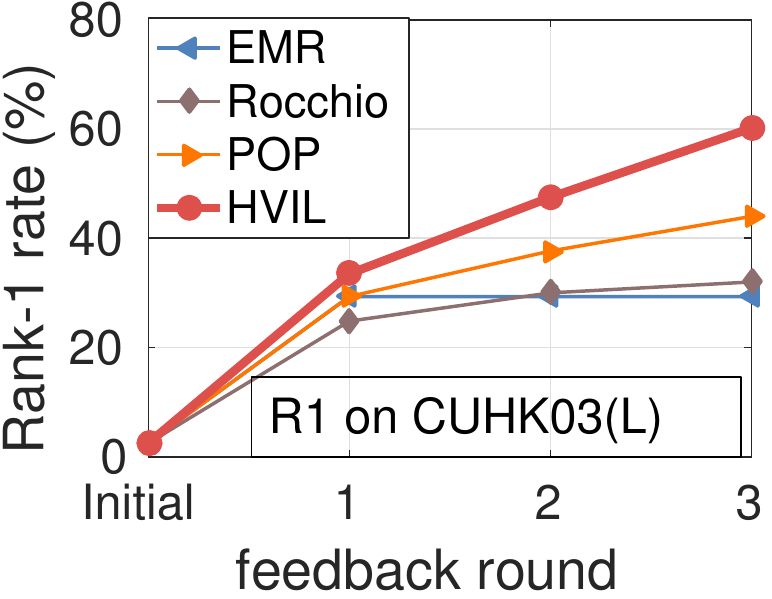}
	\includegraphics[width=0.156\textwidth]{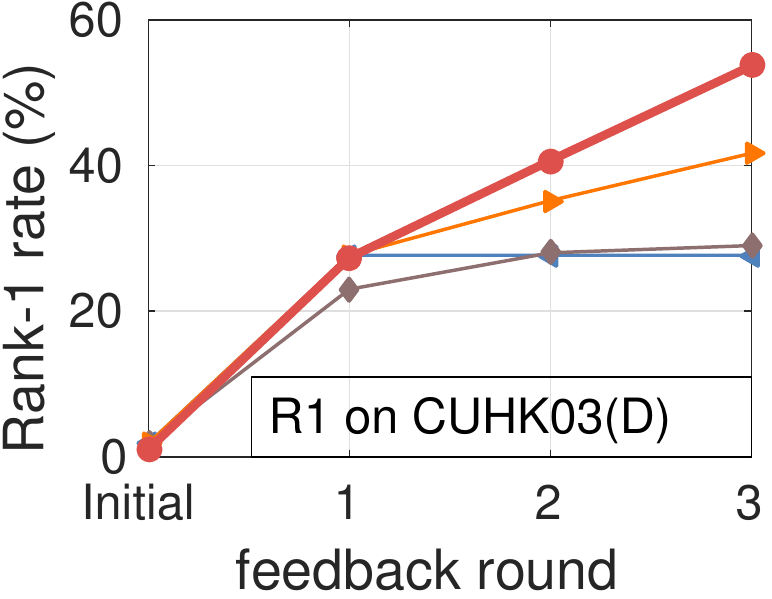}
	\includegraphics[width=0.156\textwidth]{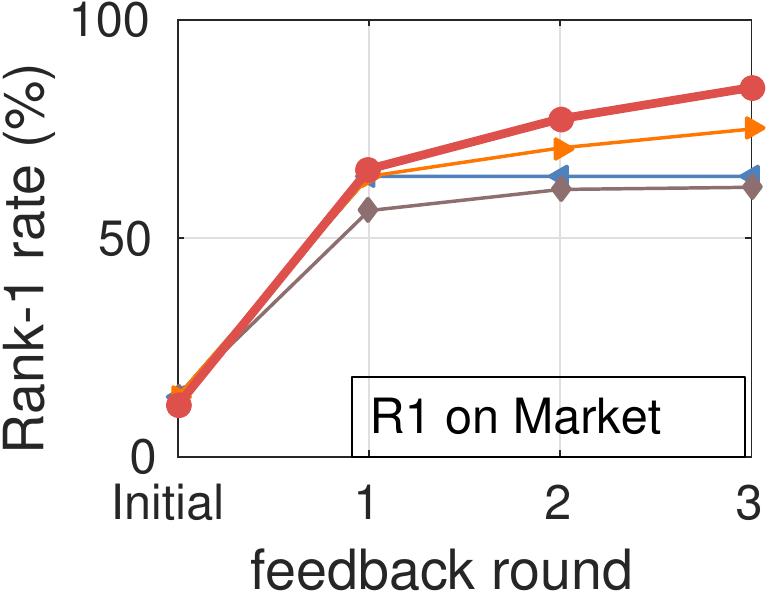}
	\includegraphics[width=0.156\textwidth]{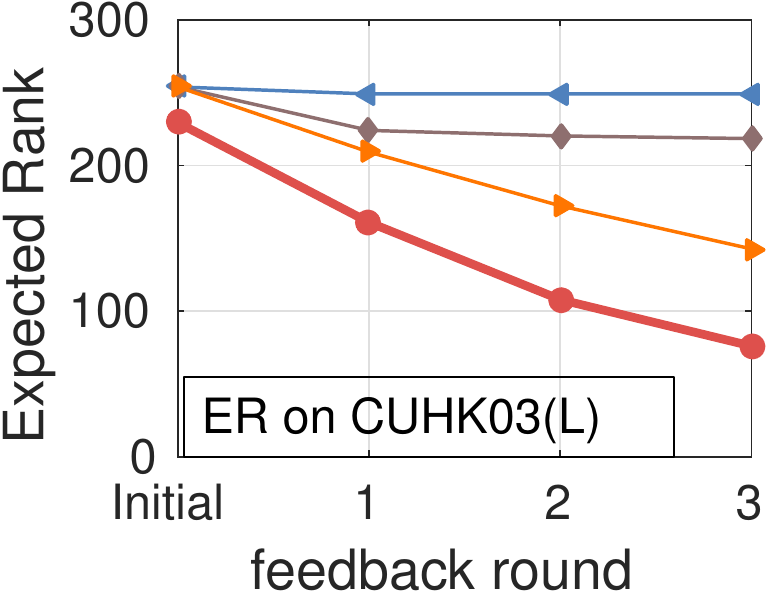}
	\includegraphics[width=0.156\textwidth]{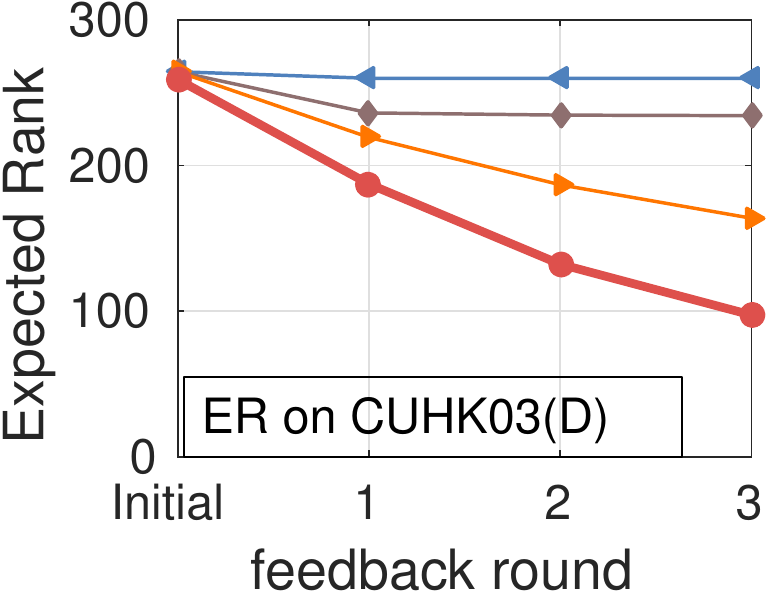}
	\includegraphics[width=0.156\textwidth]{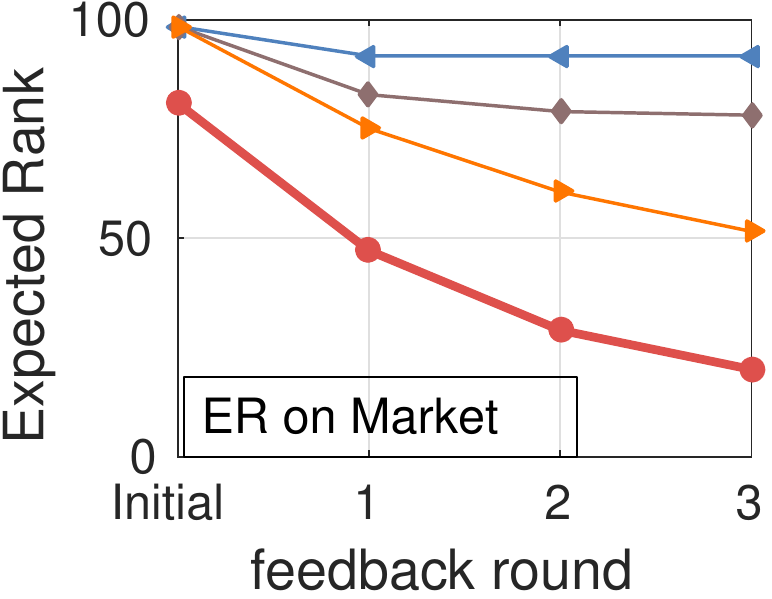}
	\vskip -0.1cm
	\caption{Comparing Rank-1 score and Expected Rank (ER) on human feedback rounds.}
	\label{fig:feedback1}
	\vspace{-0.1cm}
\end{figure}

\subsubsection{Further Analysis on Human Verification}
We examined the effectiveness of the proposed HVIL model in exploiting
human verification effort for HIL re-id 
in the single-shot setting with the WHOS feature.

\vspace{0.1cm}
\noindent \textbf{Statistics Analysis on Human Verification }
Fig.~\ref{fig:feedback1} shows the comparisons of Rank-1 and Expected 
Rank (ER) on the 4 human-in-the-loop models over three verification
feedback rounds. 
It is evident that the proposed HVIL model is more effective than the
other three models in boosting Rank-1 scores and pushing up true matches' ranking orders. 
The reasons are:
(1) Given a large gallery population with potentially complex manifold structure,
it is difficult to perform accurately graph label propagation
for graph-based methods like POP and EMR.
(2) Unlike POP/EMR/Rocchio, the proposed HVIL model optimises on  
re-id ranking losses (Eqn.~\eqref{eqn:loss})
specifically designed to maximise the two types of human verification feedback.
(3) The HVIL model enables knowledge cumulation
(Eqn.~\eqref{obj}). This is evident in Fig.~\ref{fig:feedback1} where
HVIL yields notably better (lower) Expected Ranks (ER), even for the
initial ER before verification feedback takes place on a probe
(due to benefiting cumulative effect \change{on sequential human feedback} from other probes).
In contrast, other models do not improve initial ER on each probe due
to the lack of a mechanism to cumulate experience \change{on-the-fly}.

\begin{table}[!t]
	\caption{\footnotesize
		Human verification effort
		vs. benefit. All measures are from averaging over all probes.
	Setting: single-shot. Feature: WHOS.
	$\downarrow$: lower better;
	$\uparrow$: higher better.
	\change{ES: Exhaustive Search.}
}
	\centering
	\vskip -0.3cm
		\renewcommand{\arraystretch}{1.2}
		\setlength{\tabcolsep}{0.04cm}
		\begin{tabular}{|c|ccc|ccc|ccc|}
			\hline
			Dataset   
			& \multicolumn{3}{c|}{CUHK03 (L)} 
			& \multicolumn{3}{c|}{CUHK03 (D)} 
			& \multicolumn{3}{c|}{Market-1501 (D)} \\ \hline
			Method    
			& HVIL     & POP     & ES     
			& HVIL     & POP       & ES 
			& HVIL     & POP       & ES      \\ \hline \hline
			Found-matches($\%$) $\uparrow$ 
			&  60.2  & 44.0 & \bf 100 
            & 53.7   & 41.7 & \bf 100			
			& 84.5   & 75.0  & \bf 100  \\ \hline 
			Browsed-images $\downarrow$ 
			& \bf 35.1 & 57.3  & 253.9   
            & \bf 71.6 & 107.0 & 264.3			
			& \bf 19.7 & 33.8  & 98.5    \\ \hline 
			Feedback $\downarrow$     
			& \bf 2.2   & 2.4     & -       
			& \bf 2.4   & \bf 2.4     & -   
			& \bf 1.6   & 1.7     & -        
			\\ \hline
			Search-time(sec.) $\downarrow$ 
			& \bf 23.5 &  47.3  & 187.0 
			& \bf 33.0 &  55.8  & 234.9
			& \bf 14.7 &  22.7 & 131.8
			\\ \hline
	\end{tabular}
	\label{table:statics}
	\vspace{-0.1cm}	
\end{table}

\vspace{0.1cm}
\noindent {\bf Human Verification Cost-Effectiveness }
We further evaluated the human verification effort in relation to
re-id performance benefit by analysing the meta statistics of HIL re-id
experiments above. We compared the HVIL model with
the POP model and Exhaustive Search (ES) where a user performs 
exhaustive visual searching over the whole 
gallery ranking list (1,000) generated by L2 metric until finding a true match.
The averaged statistics over all 10 trials 
were compared in Table \ref{table:statics}.
It is evident that though ES is guaranteed to 
locate a true match for every probe if it existed, 
it is much more expensive than POP ($3 \times $) and HVIL
($5\times$) in search time given a 1,000-sized gallery.
This difference will increase further on larger galleries.
Comparing HVIL and POP, it is evident that HVIL is both more
cost-effective (less Search-time, Browsed-images and Feedback) and
more accurate (more Found-matches).

\begin{figure} [!t] 
	\centering
	\includegraphics[width=0.3\textwidth]{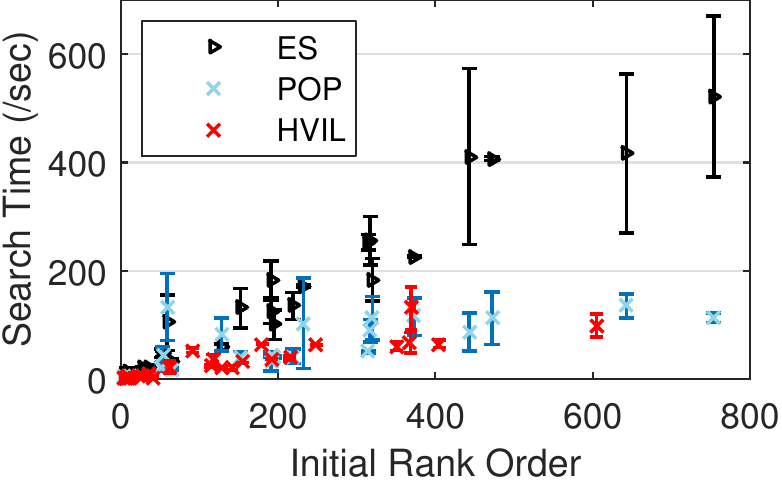}
	\vskip -0.1cm
	\caption{
		Search time from different HIL models on the same 25 randomly selected probes.
		Dataset: CUHK03 (Labelled). Setting: single-shot.
	} 
	\label{fig:time}
	\vspace{-0.1cm}
\end{figure}

\vspace{0.1cm}
\noindent {\bf HIL Re-Id Search Speed }
To better understand model convergence given human
feedback, we conducted a separate experiment to measure the search
time by different human-in-the-loop models given the initial rank 
lists on 25 randomly selected probes verified by multiple users.
This experiment was evaluated by 10 independent 
sessions with the same set of 25 probes provided.
In each session, the users were required to find a true match for
all 25 probes.
Specifically, for HVIL and POP, if a true match was
not identified after 3 (maximum) feedback,
the users then performed an exhaustive searching until it was found.
The search time statistics for all 25 probes are shown in
Fig.~\ref{fig:time}, where a bar shows the variance between 10 different sessions.
It is unsurprising that ES is the least efficient whilst HVIL is the
quickest in finding a true match, i.e. the data points of HVIL are
much lower in search time. 
Moreover, it is evident that HVIL yields much better initial 
ranks, 
i.e. the data points of HVIL are more centred towards the bottom-left corner.  
This further shows the benefit of cumulative learning in HVIL (Sec.~\ref{sec:model_update}).

\begin{table} [h]
\caption{Effect of strong and weak negatives in HIL re-id performance.}
\centering
\renewcommand{\arraystretch}{1.1}
\setlength{\tabcolsep}{0.2cm}
\label{tab:strong_weak}
\vskip -0.3cm
\begin{tabular}{|c|cc|cc|cc|}
\hline
Dataset & \multicolumn{2}{c|}{CUHK03 (L)}  & \multicolumn{2}{c|}{CUHK03 (D)}  & \multicolumn{2}{c|}{Market-1501 (D)} \\ \hline
Metric & R1(\%) & ER & R1(\%) & ER & R1(\%) & ER \\ \hline \hline 
Strong & \bf 60.2 & \bf 76.0 & \bf 53.7 & \bf 99.8 & \bf 84.5 & \bf 20.0
\\ 
Weak   & 45.3 & 203.0 & 43.6 & 226.7 & 78.0 & 90.7
\\ \hline
\end{tabular}
\end{table}

\vspace{0.1cm}
\noindent {\bf Strong vs. Weak Negatives } 
We evaluated explicitly the effect of strong and weak negative
feedback on the HIL re-id performance. 
To this end, a further experiment was conducted with the single-shot gallery setting with the WHOS feature, under the same human feedback protocol as described in Sec.~\ref{sec:human_protocol} with the only difference that users were required to label visually similar samples (weak negative) instead of dissimilar ones (strong negative). For model updates on weak negatives we adopted the same loss design of our preliminary model~\cite{wang2016human}. Table~\ref{tab:strong_weak} shows that labelling weak negatives is much less effective than strong ones
in re-id performance. For example, when weak negatives are labelled instead of strong ones,
Rank-1 rates drop from $60.2\%/53.7\%/84.5\%$ to $45.3\%/43.6\%/78.0\%$ 
and Expected Ranks increase from $76.0/99.8/20.0$ to $203.0/226.7/90.7$ on CUHK03(Labelled/Detected) and Market-1501. Moreover, it is indicated by the users that weak negatives are much harder and time consuming to label. This is intuitive given that most top-ranked gallery images are visually similar which renders a user hard to select a specific one against the others (Fig. \ref{fig:model_flow}(c)).

\subsection{Human-Out-of-the-Loop Re-Id Evaluations}
\label{sec:eval_auto_reid}
\subsubsection{Experiment Settings}
\label{sec:eval_auto_reid_set}
Finally, we assume that a limited budget for human verification
on $D_{p1}^i$ has been reached after time $\tau$ so that human feedback becomes unavailable.
Re-id of any further independent population 
(e.g. $D_{p2}^i$) turns to a conventional 
human-out-of-the-loop (HOL) re-id problem, 
if one treats previously human labelled samples  as training samples.
The proposed RMEL model was then evaluated
under this HOL re-id setting against both state-of-the-art 
supervised models and baseline ensemble models.
This experiment was conducted with CNN feature
on both CUHK03~(Labelled) and Market-1501 dataset.
Additionally, to examine our proposed HVIL-RMEL framework 
in a more comparable context defined in the
literature on HOL re-id, we also tested
on the VIPeR~\cite{VIPeR} benchmark, with more details as follows.

\vspace{0.1cm}
\noindent {\bf Training/Testing protocol } 
On CUHK03 and Market-1501
datasets, in each of the overall 10 trials,
we employed the human verified true matches
on $D_{p1}^i$ 
to learn the weights for constructing a strong ensemble model using all the
verified weak models $\{\bm{M}_j\}_{j=1}^\tau$ collected from our
previous experiments on human-in-the-loop re-id. 
The strong ensemble model was then deployed for testing on the
separate partition $D_{p2}^i$ with the size of 360 and 501 persons for
CUHK03 and Market-1501 respectively. For performance evaluation, we 
adopted the standard single-shot test setting, i.e. randomly sampling 360
cross-camera person image pairs from CUHK03 and 501 pairs from Market-1501
on $\{D_{p2}^i\}_{i=1}^{10}$ to construct the test gallery and probe
sets over ten trials.  
On VIPeR dataset, we followed 
the exact setting of the established protocol in existing literature:
splitting the 632 identities into $50\!-\!50\%$ partitions for
training  and testing sets. For obtaining weak 
re-id models, we simulated HVIL feedback update by simply giving only
groundtrue matching pairs \change{but no} strong negatives
(Eqn.~\eqref{update}); therefore each weak model was obtained by a
true-match, using the same information as training a conventional
supervised model.
On all three datasets the averaged CMC performance over all 10 trials was compared.

\begin{table} [t] 
	\centering
	\caption{Automatic person re-id (HOL) with CMC performances on
          CUHK03 and Market-1501. Gallery Size: 360 for CUHK03 and 501
          for Market-1501\protect\footnotemark.}
	\label{tab:auto_cmc_cumar}
	\vskip -0.3cm
		\renewcommand{\arraystretch}{1.07}
		\setlength{\tabcolsep}{0.15cm}
		\begin{tabular}{|l|cccc|cccc|}
			\hline
			Dataset     
			& \multicolumn{4}{c|}{CUHK03 ($N_g = 360$)}  
			& \multicolumn{4}{c|}{Market-1501 ($N_g = 501$)} 
			\\ \hline
			Rank (\%)        
			& 1    & 5     & 10  &20   
			& 1      & 5   & 10  &20     \\ \hline \hline
			kLFDA \cite{xiong2014person}      
			& 20.6  & 43.1  & 55.8  & 67.8
			& 57.0  & 83.9  & 91.9  & \bf 96.9\\ 
			XQDA \cite{liao2015person}       
			& 19.7  & 43.6  & 56.7  & 68.9
			& 52.9  & 83.5  & 89.9  & 96.1\\ 
			MLAPG \cite{Liao_2015_ICCV}       
			& 15.8  & 35.8  & 45.6  & 57.7
			& 52.2  & 78.6   & 87.7 & 94.1\\
			NFST  \cite{zhang2016learning}
			& 22.8 & 43.1 & 56.1 & 63.7
            & 58.6 & 84.1 & 90.7 & 96.3
            \\
			HER \cite{wang2016highly}
			& \bf 25.3 & 43.3 & 55.8 & 67.1
			& 60.6 & 83.9 & 90.7 & 96.8
			\\ 
			Inception-V3 \cite{szegedy2016rethinking}
			& 18.3 & 37.8 & 50.0 & 63.1
			& 56.2 & 81.7 & 88.5 & 93.6		
			\\	
			\hline
			HVIL - $\bm{M}_{avg}$
			& 19.7  & 39.2  & 55.3   & 70.3
			& 57.3  & 85.5   & 93.0  & 96.5  \\ 
			HVIL - $\bm{M}_\tau$ 
			& 20.3  &  43.3  & 56.4   & 66.1
			& 59.3  & 86.8   & \bf 93.6  & 96.5\\ \hline \hline
			HVIL - RMEL
			& 21.9 & \bf 46.7 & \bf 59.2  & \bf 71.4
			& \bf 62.6 & \bf 87.0     &  92.3   & 96.3
			\\ \hline
	\end{tabular}
\vspace{-0.1cm}
\end{table}

\begin{table} [t] 
	\centering
	\caption{Automatic person re-id (HOL) with CMC performances on VIPeR. }
	\label{tab:auto_cmc_viper}
	\vskip -0.3cm
		\renewcommand{\arraystretch}{1.07}
		\setlength{\tabcolsep}{0.4cm}
		\begin{tabular}{|l|cccc|}
			\hline
			Dataset     
			& \multicolumn{4}{c|}{VIPeR ($N_g=316$)} \\ \hline
			Rank (\%)        
			& 1     & 5    & 10   & 20   \\ \hline \hline
		MLF~\cite{Zhao_MidLevel_2014a}& 29.1 & 52.3 & 66.0  & 79.9  \\ 
		kLFDA~\cite{xiong2014person}& 38.6 & 69.2 & 80.4  & 89.2 \\ 
		SCNCD \cite{Stan14ColorName} & 33.7 & 62.7 & 74.8 & 85.0 \\
		XQDA~\cite{liao2015person}& 40.0 & 68.1 & 80.5  & 91.1 \\ 
		MLAPG~\cite{Liao_2015_ICCV}& 40.7 & 69.9 & 82.3  & 92.4 \\ 
		RKSL~\cite{wangtowards} & 40.2 & 74.5 & \bf 85.7 & \bf 93.5 \\
		NFST~\cite{zhang2016learning}& 42.3 & 71.5 & 82.9  & 92.1\\ 
		LSSCDL \cite{LSSCDL} & 42.7 & - & 84.3 & 91.9 \\ 
		HER~\cite{wang2016highly} & 45.1 & 74.6 & 85.1 & 93.3 \\
		\hline 
		RDC-Net\cite{ding2015deep} & 40.5 & 60.8 & 70.4 & 84.4 \\
		JRL \cite{chen2016deep} & 38.4 & 69.2 & 81.3 & 90.4 \\
		DGD~\cite{xiao2016learning} & 38.6 & - & -  & - \\ 
		Gated S-CNN~\cite{Gated_SCNN} & 37.8 & 66.9 & 77.4 & - \\
		S-LSTM~\cite{S_LSTM} & 42.4 & 68.7 & 79.4 & -\\ 
		MCP~\cite{Cheng_TCP} &\bf 47.8 & \bf 74.7 & 84.8 & 91.1 \\ \hline
			HVIL - $\bm{M}_{avg}$
			& 40.8 & 66.1 & 76.9 & 86.4  \\ 
			HVIL - $\bm{M}_\tau$ 
			& 42.1  & 69.0 & 78.5 & 88.6 \\ \hline \hline
			HVIL - RMEL
		    & 47.1 & 71.7 & 82.5 & 91.3 \\ \hline
	\end{tabular}
\vspace{-0.1cm}
\end{table}

\footnotetext{In this study, a challenging single-shot
training/testing protocol (300/360 for CUHK03 and 300/501 for Market-1501)    
is adopted for HOL evaluation (Table \ref{tab:auto_cmc_cumar}). 
In contrast to the reported multi-shot
setting~\cite{Li_DeepReID_2014b,zheng2015scalable} of 1260/100 for
CUHK03 and 751/750 for Market-1501, this is a harder task.
}

\vspace{0.1cm}
\noindent {\bf HOL Competitors } 
On CUHK03 and Market-1501, 
\change{six} state-of-the-art supervised re-id models are compared
\footnote{\change{Due to small training data, 
		DGD (trained from scratch by design) runs into difficulty with converging therefore excluded in comparison, 
		whilst Inception-V3 can avoid this problem by benefiting
		from model pre-training on ImageNet.
}}:
kLFDA~\cite{xiong2014person},
XQDA~\cite{liao2015person}, 
MLAPG~\cite{Liao_2015_ICCV}, 
NFST~\cite{zhang2016learning},
HER~\cite{wang2016highly},
Inception-V3~\cite{szegedy2016rethinking}.  
\change{The five metric learning methods were trained using 300 ground-truth labelled data from
$\mathcal{P}^i$ (300) and $\mathcal{G}_s^i$ (1,000) of $D_{p1}^i$
with the same CNN feature for both datasets.
The Inception-V3 was first pre-trained on
	large ImageNet~\cite{krizhevsky2012imagenet} data then
	fine-tuned on small re-id training data.}
The trained models were tested on the 
separate partition $D_{p2}^i$ with 
same testing protocol as above.
On VIPeR, as our training/testing protocol is standard,
we compared fifteen recently published state-of-the-art
including six deep models:
RDC-Net\cite{ding2015deep},
JRL \cite{chen2016deep},
DGD~\cite{xiao2016learning},
Gated S-CNN~\cite{Gated_SCNN},
S-LSTM~\cite{S_LSTM},
MCP~\cite{Cheng_TCP},
and nine shallow models:
MLF~\cite{Zhao_MidLevel_2014a},
kLFDA~\cite{xiong2014person},
SCNCD \cite{Stan14ColorName},
XQDA~\cite{liao2015person},
MLAPG~\cite{Liao_2015_ICCV},
RKSL~\cite{wangtowards},
NFST~\cite{zhang2016learning},
LSSCDL \cite{LSSCDL},
HER~\cite{wang2016highly}.
\change{Since all these existing methods utilised
	the same training/testing protocol, 
	we directly compared ours with their reported results
	optimised by the original authors.}

\vspace{0.1cm}
\noindent {\bf Metric Ensemble Baselines } 
\change{To investigate the metric ensemble effect by RMEL,}
two baseline methods are compared:
(1) HVIL - $\bm{M}_\tau$: The incrementally optimised re-id model
$\bm{M}_\tau$ obtained by HVIL from the last probe image at time
$\tau$ during the {\em human-in-the-loop} process.
(2) HVIL - $\bm{M}_{avg}$:
A naive approach to ensemble weak models, that is, simply taking
an average weighting of all weak models $\{\bm{M}_j\}_{j=1}^\tau$  
as the ensemble re-id model. 

\subsubsection{Evaluations on Person Re-Id Performance}
Tables~\ref{tab:auto_cmc_cumar} and \ref{tab:auto_cmc_viper} report the results. 
For CUHK03, there is 
insufficient labelled data for all camera pairs during training, 
given only one pair of randomly selected single-shot images per identity.
All the models generated poor re-id performances (Rank-1 rates $< 30\%$),
much less than state-of-the-art reported in the literature. For
Market-1501, a similar problem exists although less pronounced. 
Note, the results in Table~\ref{tab:auto_cmc_cumar} are based on a single-shot
test setting. This is a much harder problem than the multi-shot test
setting \cite{zheng2015scalable} where on average 14.8 true matches exist in
the gallery for each probe. 
Given the experimental results above,
it is evident that: 
Due to (1) a larger unlabelled test gallery population
than the labelled training set, (2) a lack of sufficient multi-shot
training/testing data in many camera pairs, {\em human-in-the-loop}
approach to re-id is not only desirable, but essential for re-id in
real world applications. 

Nevertheless, for HOL re-id, the HVIL-RMEL
still achieves 
\change{competitive} performance among all the models with a Rank-1 of
$21.9\%$ (\change{$3.4\%$ lower than HER ($25.3\%$)}) on CUHK03 
and $62.6\%$ (\change{$2.0\%$ higher than HER ($60.6\%$)}) on Market-1501. 
\change{Note that, the ensemble weighting for the RMEL model is learned by 
	less true-match data (253 pairs for CUHK03 and 285 pairs for Market-1501),
as compared to more ground-truth
data (300 pairs for both benchmarks) used to train 
\change{all other alternatives}.
This observation implies that by optimising re-id in a large gallery population with human 
in the loop, 
even a HOL re-id model (e.g. HVIL-RMEL) can benefit from stronger re-id generalisation to a new gallery search population.
}
%
When HVIL-RMEL was evaluated under the
standard training/testing setting on VIPeR, it yields $47.1\%$ for Rank-1 rate,
only $0.6\%$ lower compared to the current
best deep model MCP~\cite{Cheng_TCP}.
 It is also evident that
naively taking an average ensemble model (HVIL - $\bm{M}_{avg}$) gives
even poorer performance than the cumulatively learned single
model (HVIL - $\bm{M}_{\tau}$). 


\section{Conclusion}

We formulated a novel approach to human-in-the-loop person re-id by
introducing a Human Verification Incremental Learning (HVIL)
model, designed to overcome two unrealistic assumptions
adopted by existing \change{fully automated re-id models that prevent
them from being scalable to real world applications}. In particular, the
proposed HVIL model avoids the need for collecting offline
pre-labelled training data and is scalable to re-id tasks in
large gallery sizes \change{in open world search scenarios}. 
The advantage of HVIL over other
human-in-the-loop models is its ability to learn cumulatively from
human feedback on 
\change{all probed} images when available. We further
developed a regularised metric ensemble learning 
(RMEL) method to explore HVIL for automated re-id tasks
when human feedback is unavailable. Extensive comparisons
on the CUHK03~\cite{Li_DeepReID_2014b},
Market-1501~\cite{zheng2015scalable}, 
and \change{VIPeR} \cite{VIPeR} benchmarks show the 
potentials of the proposed HVIL-RMEL model for real-world re-id deployments.
\change{We also conducted extensively component evaluation and analysis for providing 
insights into the proposed HVIL model design.}

\section*{Acknowledgements}
{
	This work was partly supported by the China Scholarship Council, Vision Semantics Ltd., 
	the Royal Society Newton Advanced Fellowship Programme (NA150459), 
	and Innovate UK Industrial Challenge Project on Developing and Commercialising Intelligent Video Analytics Solutions for Public Safety (98111-571149).
}


%

%


\ifCLASSOPTIONcaptionsoff
  \newpage
\fi


\bibliographystyle{IEEEtran}
\bibliography{HITL_ReID}

\end{document}